\documentclass[10pt,twocolumn,letterpaper]{article}

\pdfoutput=1

\usepackage{cvpr}
\usepackage{times}
\usepackage{epsfig}
\usepackage{graphicx}
\usepackage{amsmath}
\usepackage{amssymb}
\usepackage[utf8]{inputenc}

\usepackage{booktabs}
\usepackage{multirow}
\usepackage{bbm}
\usepackage{xcolor}
\usepackage{subcaption}
\usepackage{overpic}
\usepackage{pict2e}

\newcommand{\documenttitle}{Multivariate Confidence Calibration for Object Detection}

\definecolor{confidence}{rgb}{0.12, 0.46, 0.71}
\definecolor{ece}{rgb}{0.84, 0.15, 0.16}

\newcommand{\realdigits}{\mathbb{R}}

\newcommand{\pdf}{f}

\newcommand{\prob}{\mathbb{P}}
\newcommand{\probscore}{p}
\newcommand{\priormodel}{\pi}
\newcommand{\estimatedmodel}{\hat{\pi}}

\newcommand{\expectation}{\mathbb{E}}

\newcommand{\betafunc}{B}
\newcommand{\betafunca}{\alpha}
\newcommand{\betafuncb}{\beta}

\newcommand{\betaparama}{a}
\newcommand{\betaparamb}{b}
\newcommand{\betaparamc}{c}

\newcommand{\betaratio}{\lambda}

\newcommand{\meanvec}{\mu}

\newcommand{\covMatP}{ \Sigma}
\newcommand{\covMatCompound}{V}

\newcommand{\classvariate}{\mathrm{y}}
\newcommand{\predclassvariate}{\hat{\mathrm{y}}}
\newcommand{\matchedvariate}{{\mathrm{m}}}
\newcommand{\predvariate}{\hat{\mathrm{p}}}
\newcommand{\bboxvariate}{\mathrm{r}}
\newcommand{\predbboxvariate}{\hat{\mathrm{r}}}
\newcommand{\collectvariate}{\mathrm{s}}

\newcommand{\classset}{\mathcal{Y}}

\newcommand{\singleinput}{x}
\newcommand{\class}{y}
\newcommand{\predclass}{\hat{\class}}

\newcommand{\pred}{\hat{\probscore}}
\newcommand{\bbox}{r}
\newcommand{\predbbox}{\hat{\bbox}}

\newcommand{\collect}{s}

\newcommand{\numcollectedvariates}{K}
\newcommand{\indexcollectedvariates}{k}

\newcommand{\numvariates}{J}

\newcommand{\numbins}{N}
\newcommand{\indexbins}{n}
\newcommand{\bin}{I}

\newcommand{\logit}{z}

\newcommand{\loglikelihoodratio}{\ell r}

\newcommand{\trainingsset}{\mathcal{D}}

\newcommand{\model}{h}
\newcommand{\calmodel}{g}

\newcommand{\confidence}{\hat{p}}
\newcommand{\calibrated}{\hat{q}}

\newcommand{\centerx}{c_x}
\newcommand{\centery}{c_y}
\newcommand{\width}{w}
\newcommand{\height}{h}

\newcommand{\scalevec}{w}
\newcommand{\scalebias}{c}

\graphicspath{{img/}}

\usepackage[breaklinks=true,bookmarks=false, pdftitle={\documenttitle}]{hyperref}

\cvprfinalcopy 

\def\cvprPaperID{5} 

\begin{document}
	
	\title{\documenttitle}
	
	\author{Fabian Küppers\textsuperscript{1}, Jan Kronenberger\textsuperscript{1}, Amirhossein Shantia\textsuperscript{2}, Anselm Haselhoff\textsuperscript{1}\\
	{\tt\small \{fabian.kueppers, jan.kronenberger, anselm.haselhoff\}@hs-ruhrwest.de, ashantia@visteon.com}\\
	\and
	\textsuperscript{1}Ruhr West University of Applied Sciences\\
	Bottrop, Germany
	\and
	\textsuperscript{2}Visteon Electronics GmbH\\
	Kerpen, Germany
	}
	
	\maketitle
	\thispagestyle{empty}
	
	\begin{abstract}
	Unbiased confidence estimates of neural networks are crucial especially for safety-critical applications. Many methods have been developed to calibrate biased confidence estimates. Though there is a variety of methods for classification, the field of object detection has not been addressed yet.
	Therefore, we present a novel framework to measure and calibrate biased (or miscalibrated) confidence estimates of object detection methods \footnote{The framework is publicly available at: \url{https://github.com/fabiankueppers/calibration-framework}}. The main difference to related work in the field of classifier calibration is that we also use additional information of the regression output of an object detector for calibration.
	Our approach allows, for the first time, to obtain calibrated confidence estimates with respect to image location and box scale. In addition, we propose a new measure to evaluate miscalibration of object detectors. Finally, we show that our developed methods outperform state-of-the-art calibration models for the task of object detection and provides reliable confidence estimates across different locations and scales. 
\end{abstract}
	
	\section{Introduction}

Modern object detection algorithms are widely used in several domains like autonomous driving or medical diagnosis. Most of these object detectors, based on neural networks, provide a score for a certain class and a proposal of the object location and scale. In previous work it has been shown that modern architectures for classification tasks tend to be overconfident with their predictions compared to the observed accuracy \cite{Guo2018} and thus are called \textit{miscalibrated}. Calibration in the scope of predictors means that the observed accuracy of the predictor matches its mean confidence. Miscalibration can be observed for modern object detection models as well. However, a well-calibrated object detector is crucial especially in safety-critical applications like pedestrian detection for autonomous driving applications.

For image classification, there are methods to calibrate the predictions of a network as a post-processing step (logistic calibration/Platt scaling \cite{Platt1999}, temperature scaling \cite{Guo2018}, etc.). These methods can easily be applied to object detection tasks.
\begin{figure}[t]
	\centering
	\begin{overpic}[width=1.0\linewidth]{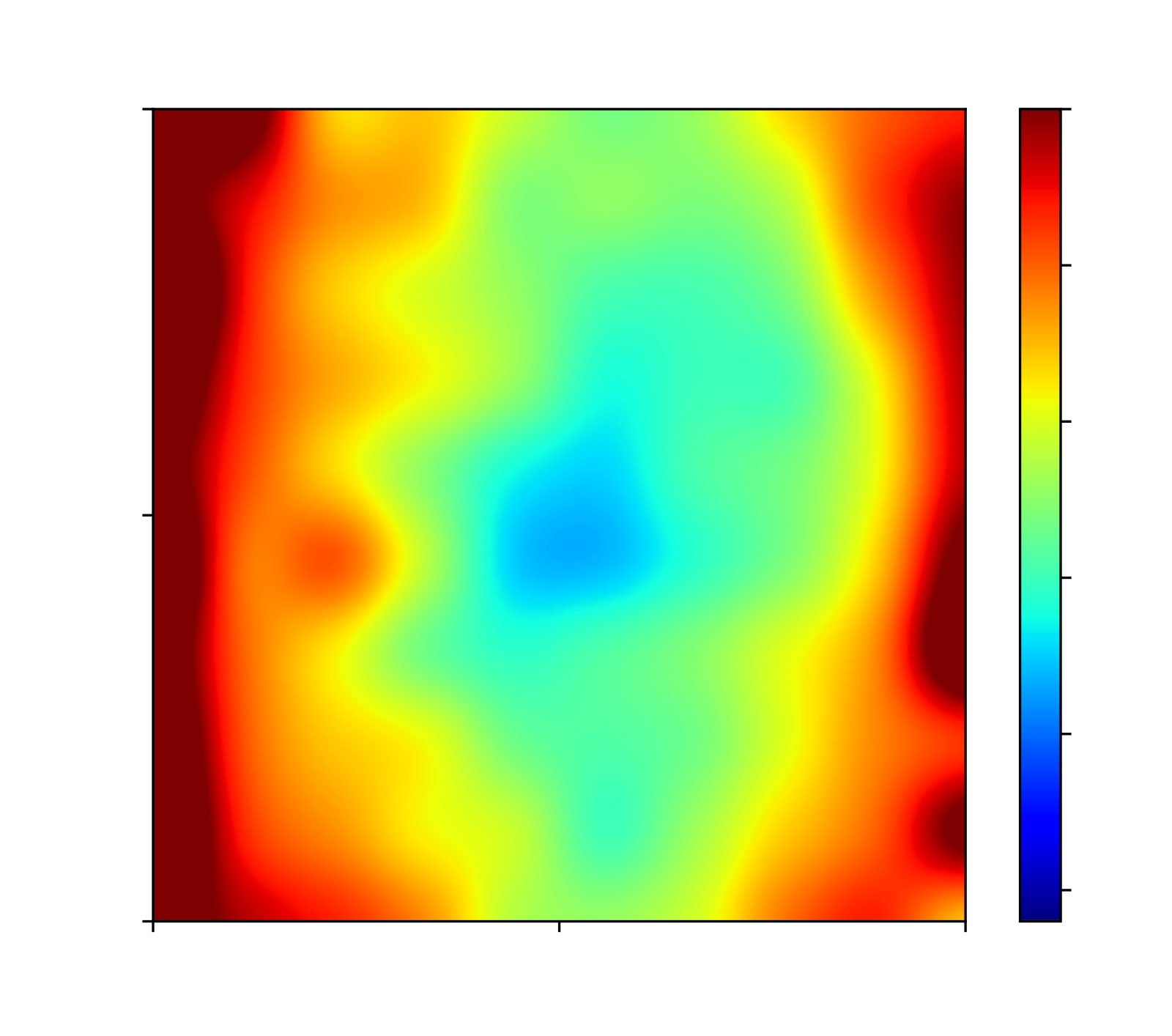}
		\put(29,85){{Miscalibration Score on}}
		\put(34,81){{COCO detections}}
		\put(0, 37){\small{\rotatebox{90}{relative $\centery$}}}
		\put(7, 8.7){\scriptsize{0.0}}
		\put(7, 43.4){\scriptsize{0.5}}
		\put(7, 77.8){\scriptsize{1.0}}
		
		\put(42, -1){\small{relative $\centerx$}}
		\put(11, 5){\scriptsize{0.0}}
		\put(46, 5){\scriptsize{0.5}}
		\put(81, 5){\scriptsize{1.0}}
		
		\put(94, 11.3){\scriptsize{5}}
		\put(93, 24.6){\scriptsize{10}}
		\put(93, 38.2){\scriptsize{15}}
		\put(93, 51.4){\scriptsize{20}}
		\put(93, 64.8){\scriptsize{25}}
		\put(93, 77.7){\scriptsize{30}}
		\put(83, 81){\small{D-ECE [\%]}}
	\end{overpic}
	\caption{Miscalibration evaluation on the COCO dataset for a R-FCN ResNet-101 \cite{Dai2016}. The miscalibration score highly depends on the object location (the center coordinates $\centerx$ and $\centery$) and increases as the prediction gets close to the image boundaries. The evaluation is based on the detection expected calibration error (D-ECE), a variant of the expected calibration error (ECE) \cite{Naeini2015}.}
		\label{img:intro:miscalibration}
\end{figure}
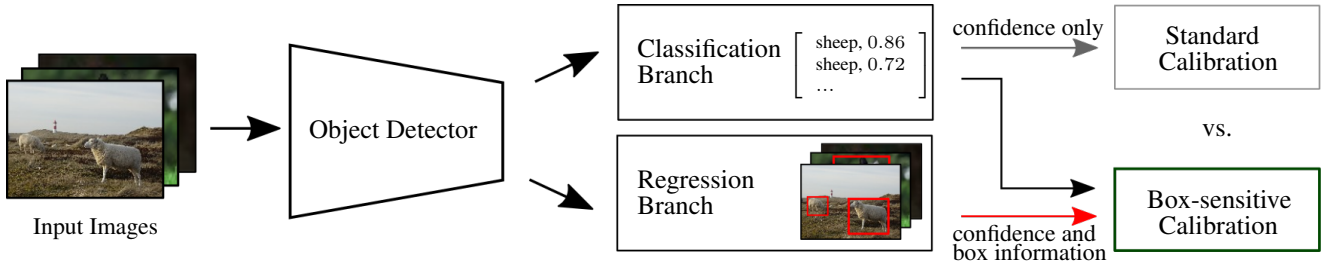
\begin{figure*}[ht]
	\centering
	\begin{overpic}[width=1.0\linewidth]{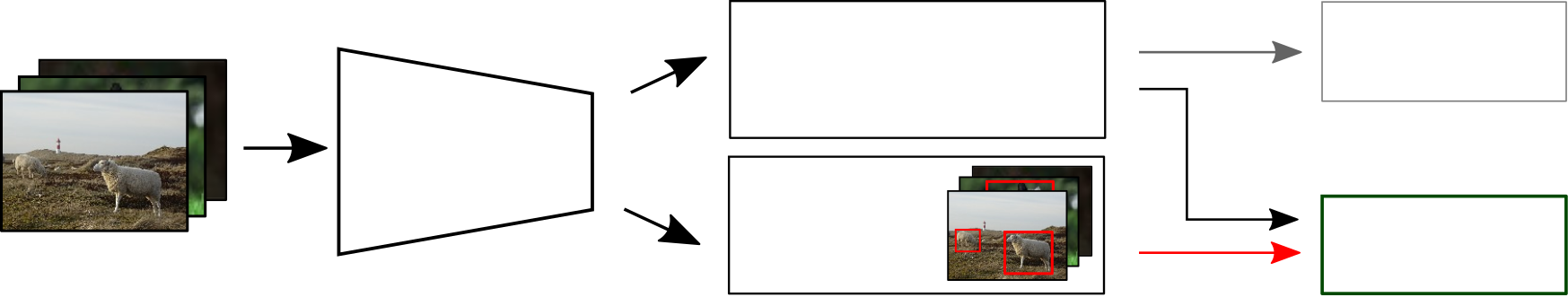}
		\put(2, 2){\small{Input Images}}
		
		\put(23, 9.3){Object Detector}
		\put(48, 15.5){Classification}
		\put(48, 13.5){Branch}
		\put(59.5, 14.5){\scriptsize{$\left[\begin{array}{l}
			\text{sheep, } 0.86 \\
			\text{sheep, } 0.72 \\
			...  \\
			\end{array}\right]$}}
		
		\put(48, 5.5){Regression}
		\put(48, 3.5){Branch}
		
		\put(72, 17.2){\small{confidence only}}
		\put(72, 1.5){\small{confidence and}}
		\put(72, 0.0){\small{box information}}
		
		\put(88.2, 16.5){Standard}
		\put(87.7, 14.5){Calibration}
		\put(91, 9.5){vs.}
		\put(86.8, 4){Box-sensitive}
		\put(87.7, 2){Calibration}
	\end{overpic}
	\caption{Extended box-sensitive or multivariate calibration proposed in this paper (bottom) vs. standard confidence calibration (top). An object detector predicts a class with a certain confidence and a box proposal containing information about location and scale. Standard calibration methods only take the confidence into account. Our approach also includes the regression branch of an object detector in order to improve the calibration results.}
	\label{img:intro:flowdiagram}
\end{figure*}
However, these well-known methods only take the confidence scores of the detections into account.
As illustrated in Fig.~\ref{img:intro:miscalibration}, the calibration error highly depends on the object location and increases as the prediction gets close to the image boundaries. A similar behavior can be observed with respect to the object width and height. These location-dependencies are not captured by conventional calibration methods but might be very important especially in safety-critical domains. For example, in road traffic it often occurs that a pedestrian suddenly appears at the roadside to cross the road and is therefore located at the image boundaries. If we observe a miscalibration especially at the boundaries, we cannot trust in the confidence score. 

Motivated by the empirical observation that the calibration error depends on bounding box properties, our goal is to develop box-sensitive calibration methods for detectors.
We propose a simple and effective framework to measure and optimize the calibration properties of object detectors. The approach treats the detector as a black-box and is therefore applied after the regular training process. Our framework is agnostic to the used detection method as well as to the underlying calibration algorithm. In other words, we do not modify the learning objective of a detector and thus our work is orthogonal to ongoing research in regularization methods for calibration \cite{Neumann2018, Seo2019, Phan2018}.
We show how to extend common calibration algorithms, \eg logistic calibration, and how to measure miscalibration location and/or scale dependent. We use pretrained state-of-the-art object detection methods (SSD \cite{Liu2016}, Faster R-CNN \cite{Ren2015} and R-FCN \cite{Dai2016}) to evaluate our considerations about the developed calibration algorithms.

	\section{Related Work}

Common methods for confidence calibration in the domain of classification are histogram binning \cite{Zadrozny2001}, isotonic regression \cite{Zadrozny2002}, Bayesian binning into quantiles (BBQ) \cite{Naeini2015}, ensemble of near isotonic regression (ENIR) \cite{Naeini2015a}, logistic calibration/Platt scaling \cite{Platt1999}, temperature scaling \cite{Guo2018} and beta calibration \cite{Kull2017}. These methods are applied during inference as a post-processing step after classification. A straight forward transfer of these methods to object detectors is possible, but might not utilize the full potential - dependencies between the estimated confidences and the regression output might have an influence.

In contrast to post-processing calibration methods, \cite{Seo2019} has recently proposed a method to obtain well-calibrated classifiers during model training. The authors have shown that the uncertainty obtained by Monte-Carlo dropout is correlated with the miscalibration and hence they use the uncertainty to control an additional regularization term. This approach is related to the ongoing research in the domain of model uncertainty. A common way to obtain uncertainties are Bayesian neural networks (BNNs) \cite{Gal2016, Kendall2017}. An additional approach to obtain uncertainties for one-stage object detectors is presented by \cite{Kraus2019}. However, it has been shown that even those uncertainty estimates are biased and thus need calibration \cite{Gal2016, Kuleshov2018, Kraus2019}. In \cite{Kuleshov2018}, a method is proposed to perform regression calibration on the confidence interval obtained by a BNN. In contrast, \cite{Phan2018} use a regression calibration framework to obtain calibrated object locations. These approaches are orthogonal to ours and could be used in tandem with ours. For example, the framework proposed in this paper can either be used to integrate a dependence to the regression output or to further optimize the calibration in a post-processing step. 

Similarly, IoU-Net \cite{Jiang2018} and GIoU \cite{Rezatofighi2019} are a valuable extension to our proposed framework to demonstrate further orthogonal calibration directions. In contrast to our approach, IoU-Net focuses on fine-tuning the regression output to improve the IoU between predicted and ground-truth position. They introduce an additional location confidence (regression) in conjunction with the confidence (categorical). They even support the need for our calibration method because they show that the ’regular’ confidence is not a direct measure for misalignment. We agree that an improved localization might have an impact on the precision and thus on the confidence calibration as well, but the necessity of confidence calibration remains.

An approach to obtain well-calibrated confidence estimates for object detection is proposed in \cite{Neumann2018}. They perform a temperature scaling \cite{Guo2018} by adding an auxiliary output to the network that scales the remaining logits. Additionally, the authors propose a variation of the expected calibration error (ECE) \cite{Naeini2015} as the preferred miscalibration metric for object detection. However, neither the calibration nor the proposed variation of the ECE are explicitly designed to capture properties of the regression output. 

	\section{Methods for Confidence Calibration}
In this section we introduce our calibration methodology for object detectors. The main idea is to integrate the classification and regression branch of a detector to perform a box-sensitive calibration (Fig.~\ref{img:intro:flowdiagram}). First, we discuss some background about object detectors that is relevant for the calibration task. Afterwards, we introduce the calibration framework which is an extension to the multivariate domain to handle the regression branch. A simple artificial calibration example is provided to demonstrate the advantage of the proposed method. Finally, we show how to integrate standard methods, e.g. beta calibration, into the calibration framework.

\subsection{Object Detectors}
\label{section:definitions}
Modern object detectors typically take an image as the input $x$ and deliver a prediction in form of a class label $\class$, a confidence score $\probscore$ and a bounding box $\bbox = (\centerx, \centery, \width, \height)$, where $(\centerx, \centery)$ is the position of the box center and the scale is determined by the width and height $(\width, \height)$. We interpret these quantities as random variables that follow a joint ground-truth distribution $\priormodel(\singleinput, \class, \bbox) = \priormodel(\class, \bbox | \singleinput) \priormodel(\singleinput)$. The detector can be interpreted as function $\model(\singleinput) = (\predclass, \pred, \predbbox)$, that takes the input images and generates predictions for the class, confidence and location. The training process for fitting the model parameters is based on a probabilistic model $\estimatedmodel(\class, \bbox | \singleinput)$, where the cross-entropy between the training data $\priormodel_{data}(\singleinput, \class, \bbox)$ (empirical distribution of $\priormodel(\singleinput, \class, \bbox)$ given by the dataset $\trainingsset$) and the model $\estimatedmodel(\class, \bbox | \singleinput)$ is minimized. The evaluation of a detector differs from classification in the sense that both, the class label and the regression output for the location, must match the ground-truth. For the bounding box prediction $\predbbox$ it is nearly impossible to match the ground-truth perfectly, therefore, a certain overlap score (e.g. Intersection of Union (IoU)) is required to assign detections to ground-truth annotations. In other words, the class prediction and a predefined threshold, e.g. IoU $> 0.5$, is used to classify whether a prediction is a true or false positive. 

The confidence score of a regular classifier is calibrated with respect to the accuracy. For  black-box detectors, e.g. without any information about anchors, rejected proposals, etc., it is impossible to determine the accuracy. Instead, the common metrics precision and recall are used to evaluate the performance of an object detector. Precision is the fraction of correct detections among all detections while recall denotes the proportion of actual objects that are correctly identified as such among all ground-truth objects. Common methods like temperature scaling perform a logistic fit between the estimated confidence and the corresponding ground-truth label. If we want to calibrate with respect to the recall, we also need to integrate ground-truth objects into the calibration that are not identified as such and thus have no confidence score. Therefore, a calibration with respect to the recall is not possible in a black-box setting. In this work we thus use precision as a surrogate for accuracy. That is to say, given 100 detections, each predicted with 0.9 confidence, 90 out of the 100 detections should be correctly classified. This is our measure to define calibration for black-box classifiers.
Thus, an object detection model is perfectly calibrated if 
\begin{align}
\label{eq:calibration}
& \underbrace{\prob(\matchedvariate=1 | \predvariate = \probscore, \predclassvariate = \class, \predbboxvariate = \bbox)}_{\text{precision given } \probscore, \class, \bbox} = \underbrace{\probscore}_{\text{confidence}} \\ \nonumber
& \forall \probscore \in [0,1], \class \in \classset, \bbox \in [0,1]^\numvariates 
\end{align}
is fulfilled, where $\matchedvariate=1$ denotes a correctly classified prediction that matches a ground-truth object and $\matchedvariate=0$ denotes a mismatch. Thus,  $\prob(\matchedvariate=1)$ is the shorthand notation for approximating $\prob( \predclassvariate = \classvariate, \predbboxvariate = \bboxvariate)$ with a certain IoU threshold. We use bounding box information relative to the image size $\bbox \in [0,1]^\numvariates$, where $\numvariates$ is the dimension of the used box encoding. Similar to \cite{Guo2018}, we assume that a perfect calibration is not achievable by any known detection method.

\subsection{Multivariate Calibration Framework}

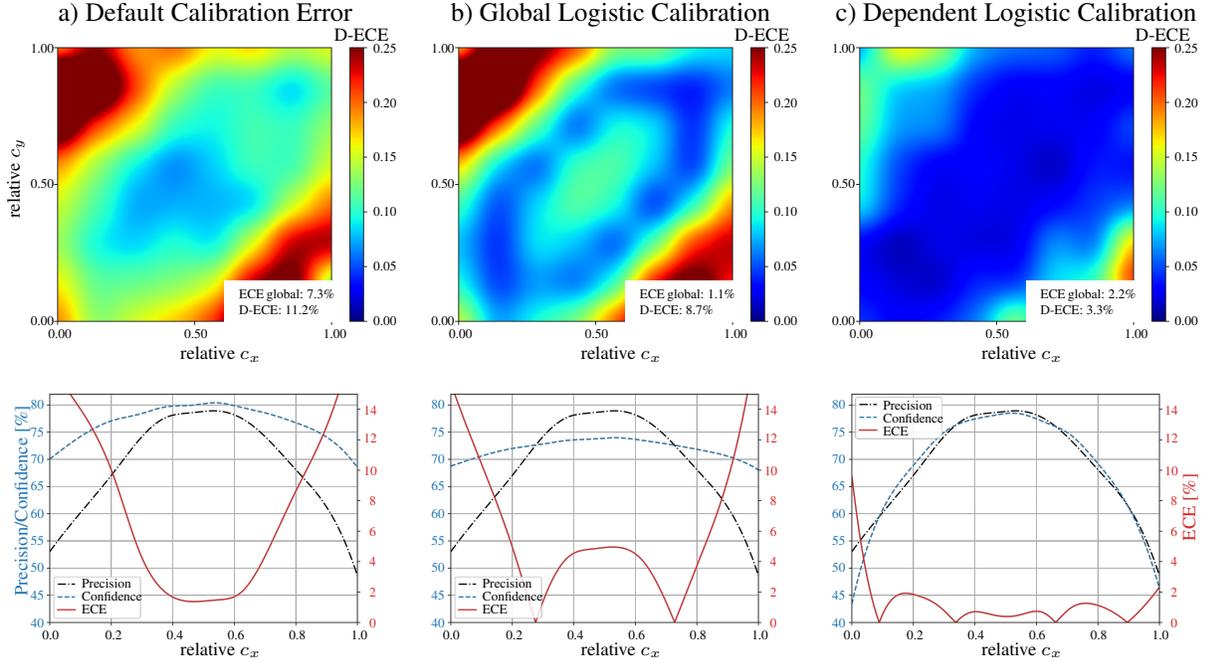
\begin{figure*}[t]
	\centering
	\begin{overpic}[width=0.3\textwidth]{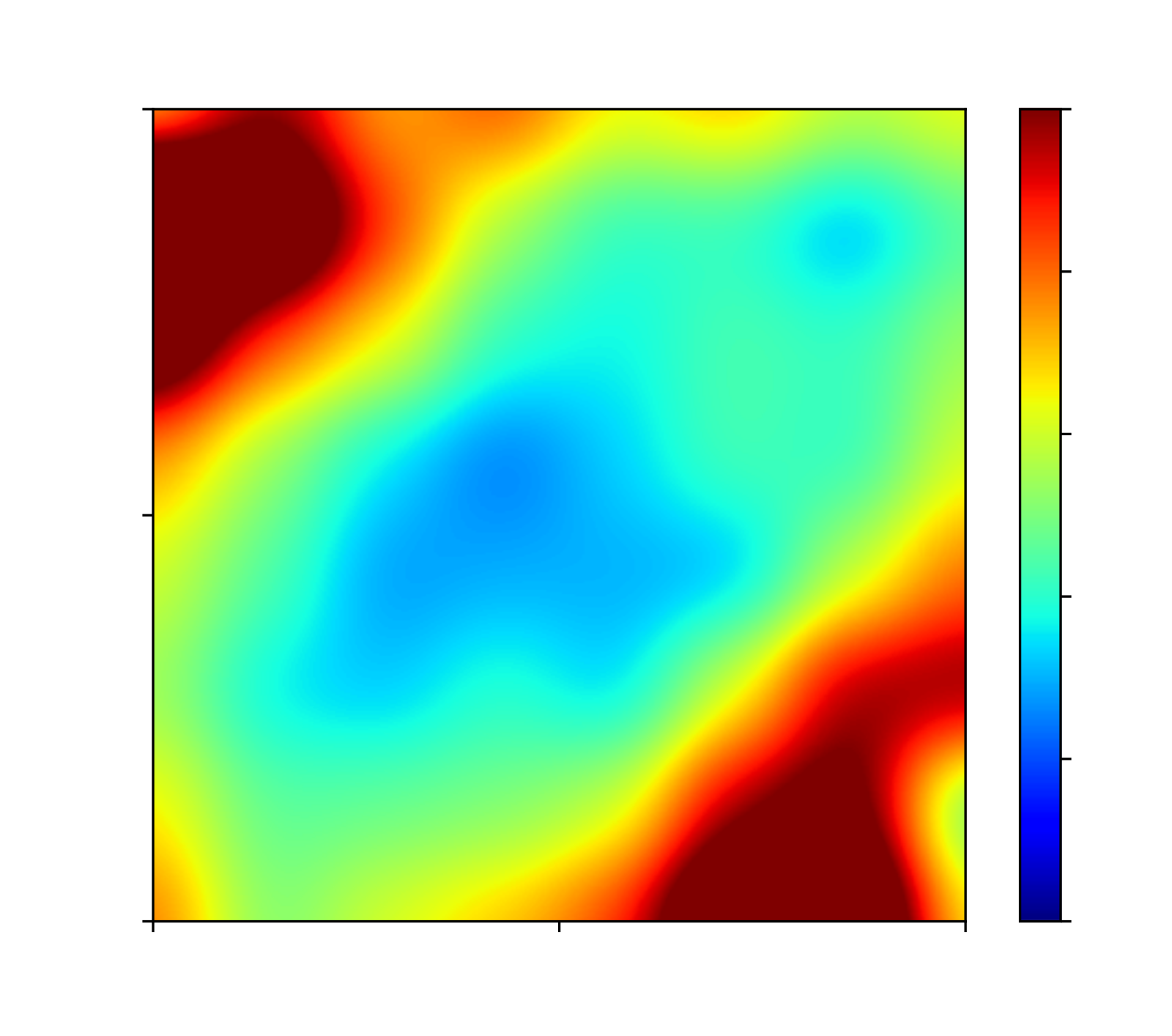}
		\put(13,86){a) Default Calibration Error}
		\put(0, 36){\rotatebox{90}{\scriptsize{relative $\centery$}}}
		\put(6, 8.6){\tiny{0.00}}
		\put(6, 43.5){\tiny{0.50}}
		\put(6, 78){\tiny{1.00}}
		
		\put(44, 0){\scriptsize{relative $\centerx$}}
		\put(10.5, 5.5){\tiny{0.00}}
		\put(45, 5.5){\tiny{0.50}}
		\put(80.3, 5.5){\tiny{1.00}}
		
		\put(83, 81){\scriptsize{D-ECE}}
		\put(93, 8.8){\tiny{0.00}}
		\put(93, 22.6){\tiny{0.05}}
		\put(93, 36.4){\tiny{0.10}}
		\put(93, 50.3){\tiny{0.15}}
		\put(93, 64.1){\tiny{0.20}}
		\put(93, 77.9){\tiny{0.25}}
		
		\put(53.55, 13.04){
			\colorbox{white}{
				\begin{minipage}[l]{3.61em}
					\tiny{ECE global: 7.3\%}\\
					\tiny{D-ECE: 11.2\%}
				\end{minipage}}
			}
	\end{overpic}
	\vspace{0.2em}
	\begin{overpic}[width=0.3\textwidth]{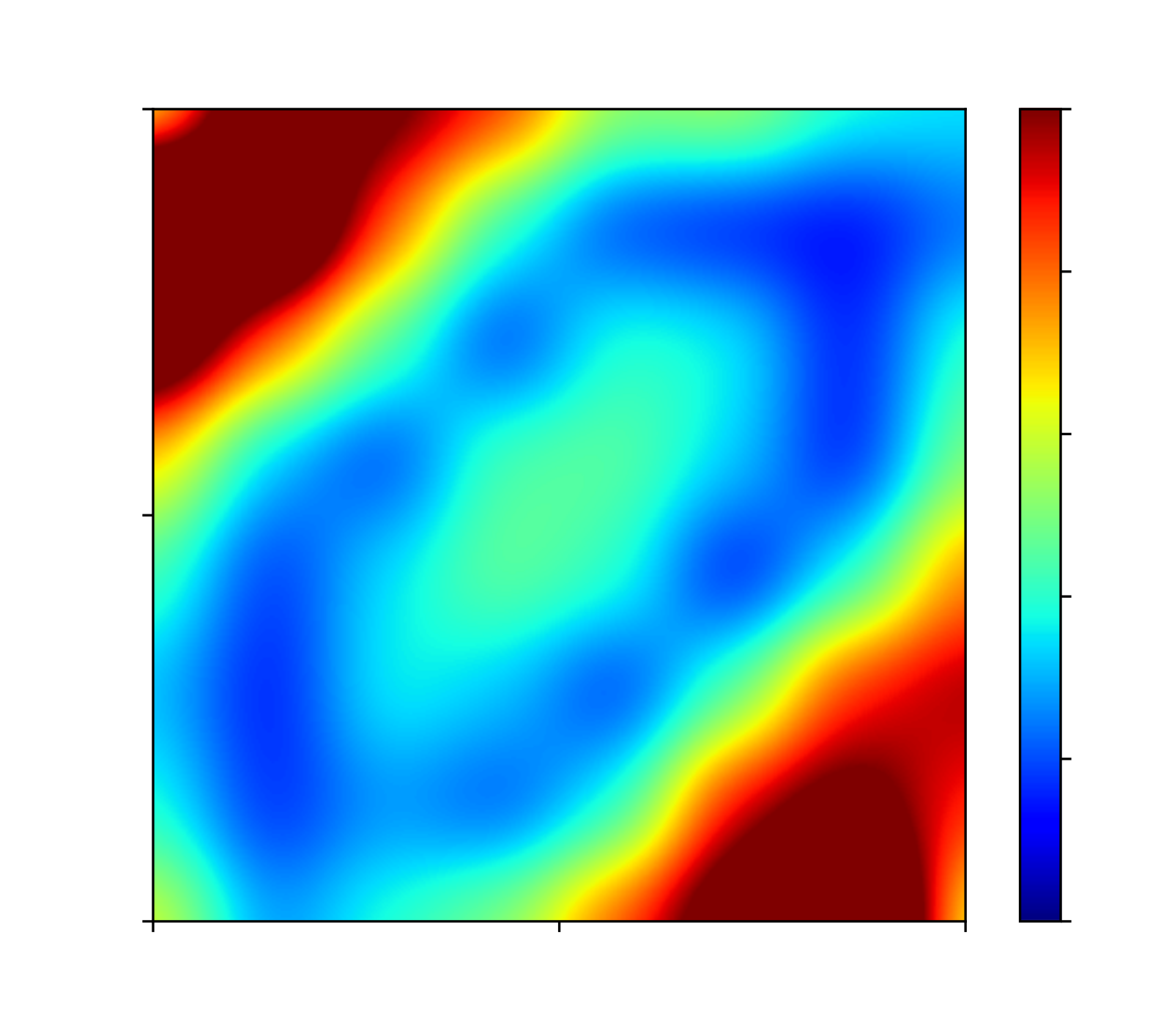}
		\put(11,86){b) Global Logistic Calibration}
		\put(6, 8.6){\tiny{0.00}}
		\put(6, 43.5){\tiny{0.50}}
		\put(6, 78){\tiny{1.00}}
		
		\put(44, 0){\scriptsize{relative $\centerx$}}
		\put(10.5, 5.5){\tiny{0.00}}
		\put(45, 5.5){\tiny{0.50}}
		\put(80.3, 5.5){\tiny{1.00}}
		
		\put(83, 81){\scriptsize{D-ECE}}
		\put(93, 8.8){\tiny{0.00}}
		\put(93, 22.6){\tiny{0.05}}
		\put(93, 36.4){\tiny{0.10}}
		\put(93, 50.3){\tiny{0.15}}
		\put(93, 64.1){\tiny{0.20}}
		\put(93, 77.9){\tiny{0.25}}
		
		\put(53.55, 13.04){
			\colorbox{white}{
				\begin{minipage}[l]{3.61em}
					\tiny{ECE global: 1.1\%}\\
					\tiny{D-ECE: 8.7\%}
			\end{minipage}}
		}
	\end{overpic}
	\begin{overpic}[width=0.3\textwidth]{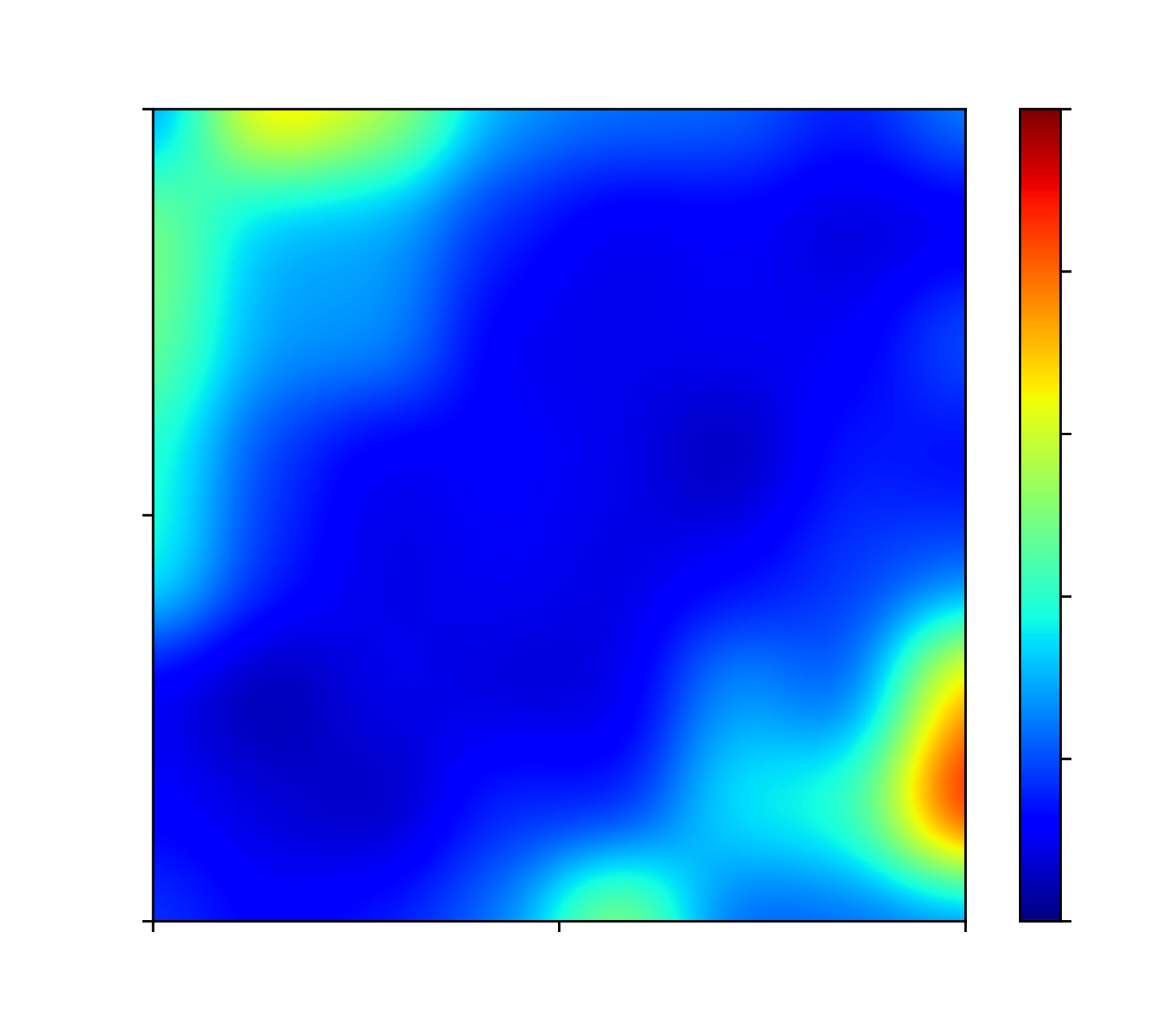}
		\put(7,86){c) Dependent Logistic Calibration}
		\put(6, 8.6){\tiny{0.00}}
		\put(6, 43.5){\tiny{0.50}}
		\put(6, 78){\tiny{1.00}}
		
		\put(44, 0){\scriptsize{relative $\centerx$}}
		\put(10.5, 5.5){\tiny{0.00}}
		\put(45, 5.5){\tiny{0.50}}
		\put(80.3, 5.5){\tiny{1.00}}
		
		\put(83, 81){\scriptsize{D-ECE}}
		\put(93, 8.8){\tiny{0.00}}
		\put(93, 22.6){\tiny{0.05}}
		\put(93, 36.4){\tiny{0.10}}
		\put(93, 50.3){\tiny{0.15}}
		\put(93, 64.1){\tiny{0.20}}
		\put(93, 77.9){\tiny{0.25}}
		
		\put(53.55, 13.04){
			\colorbox{white}{
				\begin{minipage}[l]{3.61em}
					\tiny{ECE global: 2.2\%}\\
					\tiny{D-ECE: 3.3\%}
			\end{minipage}}
		}
	\end{overpic}

	\begin{overpic}[width=0.3\textwidth]{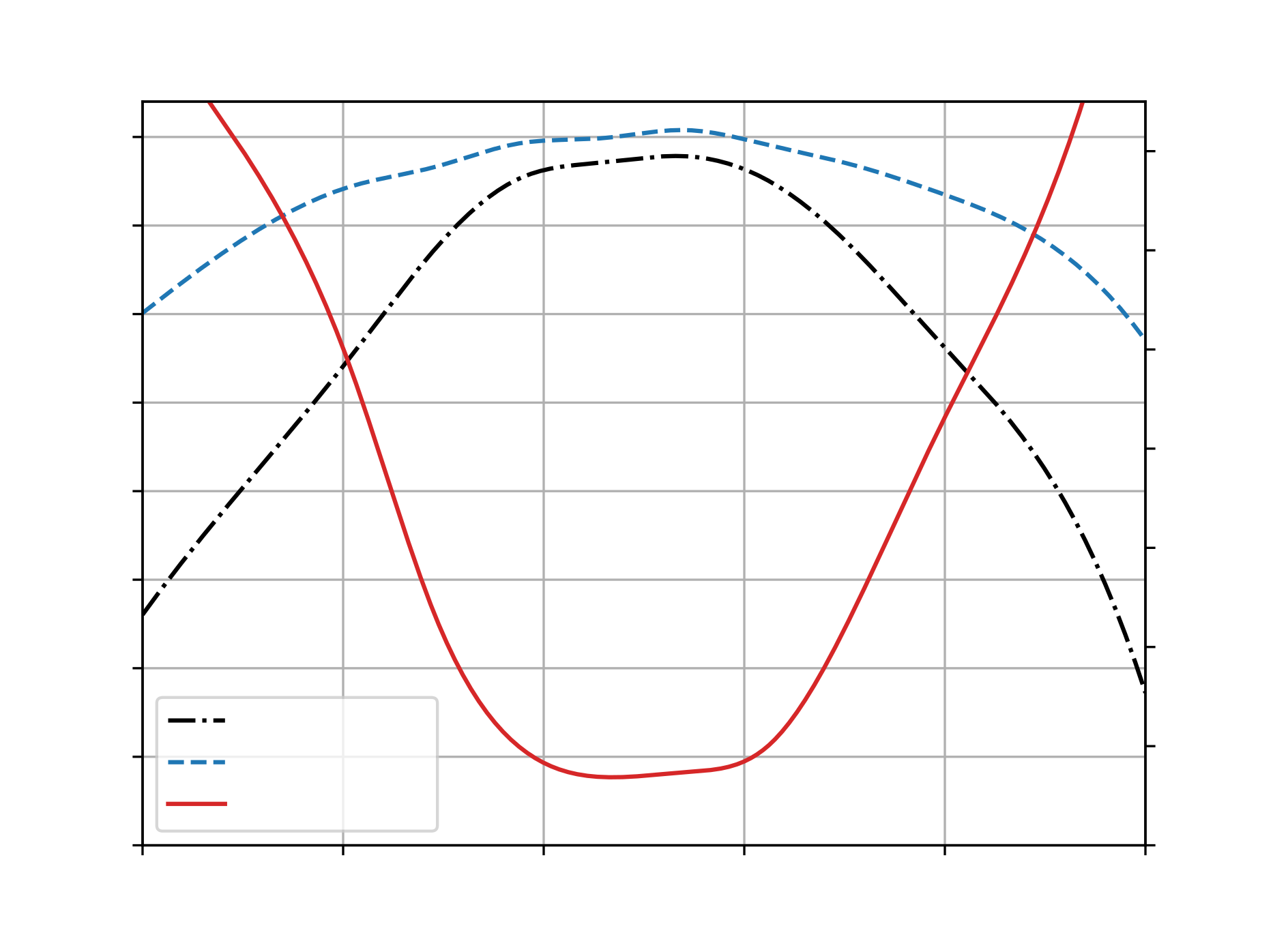}
		\put(2, 15){\rotatebox{90}{\scriptsize{\textcolor{confidence}{Precision/Confidence [\%]}}}}
		\put(6, 7){\tiny{\textcolor{confidence}{40}}}
		\put(6, 14){\tiny{\textcolor{confidence}{45}}}
		\put(6, 21){\tiny{\textcolor{confidence}{50}}}
		\put(6, 27.8){\tiny{\textcolor{confidence}{55}}}
		\put(6, 34.7){\tiny{\textcolor{confidence}{60}}}
		\put(6, 41.7){\tiny{\textcolor{confidence}{65}}}
		\put(6, 48.5){\tiny{\textcolor{confidence}{70}}}
		\put(6, 55.5){\tiny{\textcolor{confidence}{75}}}
		\put(6, 62.4){\tiny{\textcolor{confidence}{80}}}

		\put(44, 0){\scriptsize{relative $\centerx$}}
		\put(9, 4){\tiny{0.0}}
		\put(25, 4){\tiny{0.2}}
		\put(40.5, 4){\tiny{0.4}}
		\put(56, 4){\tiny{0.6}}
		\put(71.6, 4){\tiny{0.8}}
		\put(87.3, 4){\tiny{1.0}}
		
		\put(92, 7){\tiny{\textcolor{ece}{0}}}
		\put(92, 14.8){\tiny{\textcolor{ece}{2}}}
		\put(92, 22.5){\tiny{\textcolor{ece}{4}}}
		\put(92, 30.3){\tiny{\textcolor{ece}{6}}}
		\put(92, 38){\tiny{\textcolor{ece}{8}}}
		\put(90.5, 45.7){\tiny{\textcolor{ece}{10}}}
		\put(90.5, 53.8){\tiny{\textcolor{ece}{12}}}
		\put(90.5, 61){\tiny{\textcolor{ece}{14}}}
		
		\put(19, 17){\tiny{Precision}}
		\put(19, 13.7){\tiny{Confidence}}
		\put(19, 10.4){\tiny{ECE}}
	\end{overpic}
	\begin{overpic}[width=0.3\textwidth]{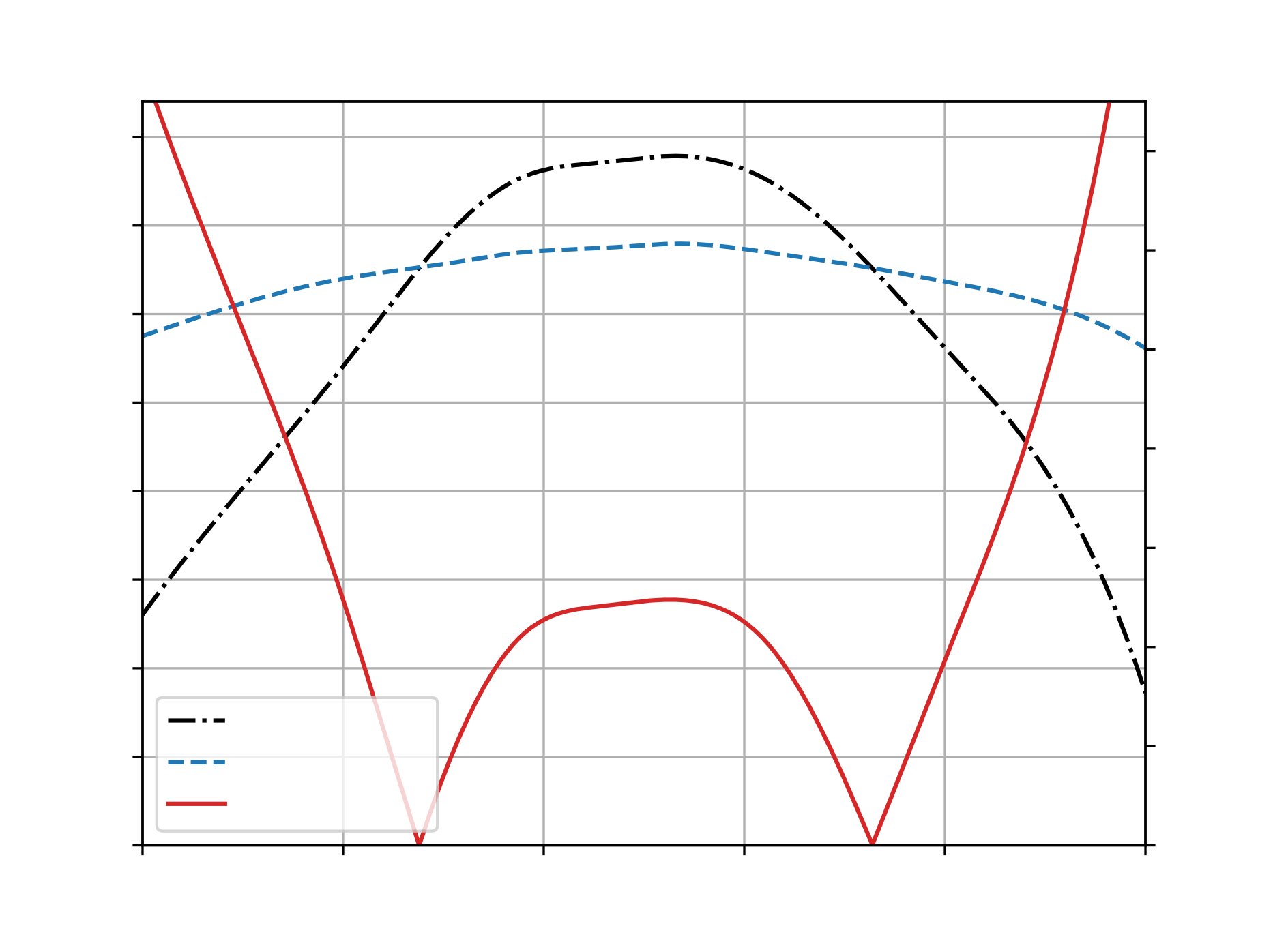}
		\put(6, 7){\tiny{\textcolor{confidence}{40}}}
		\put(6, 14){\tiny{\textcolor{confidence}{45}}}
		\put(6, 21){\tiny{\textcolor{confidence}{50}}}
		\put(6, 27.8){\tiny{\textcolor{confidence}{55}}}
		\put(6, 34.7){\tiny{\textcolor{confidence}{60}}}
		\put(6, 41.7){\tiny{\textcolor{confidence}{65}}}
		\put(6, 48.5){\tiny{\textcolor{confidence}{70}}}
		\put(6, 55.5){\tiny{\textcolor{confidence}{75}}}
		\put(6, 62.4){\tiny{\textcolor{confidence}{80}}}
		
		\put(44, 0){\scriptsize{relative $\centerx$}}
		\put(9, 4){\tiny{0.0}}
		\put(25, 4){\tiny{0.2}}
		\put(40.5, 4){\tiny{0.4}}
		\put(56, 4){\tiny{0.6}}
		\put(71.6, 4){\tiny{0.8}}
		\put(87.3, 4){\tiny{1.0}}
		
		\put(92, 7){\tiny{\textcolor{ece}{0}}}
		\put(92, 14.8){\tiny{\textcolor{ece}{2}}}
		\put(92, 22.5){\tiny{\textcolor{ece}{4}}}
		\put(92, 30.3){\tiny{\textcolor{ece}{6}}}
		\put(92, 38){\tiny{\textcolor{ece}{8}}}
		\put(90.5, 45.7){\tiny{\textcolor{ece}{10}}}
		\put(90.5, 53.8){\tiny{\textcolor{ece}{12}}}
		\put(90.5, 61){\tiny{\textcolor{ece}{14}}}
		
		\put(19, 17){\tiny{Precision}}
		\put(19, 13.7){\tiny{Confidence}}
		\put(19, 10.4){\tiny{ECE}}
	\end{overpic}
	\begin{overpic}[width=0.3\textwidth]{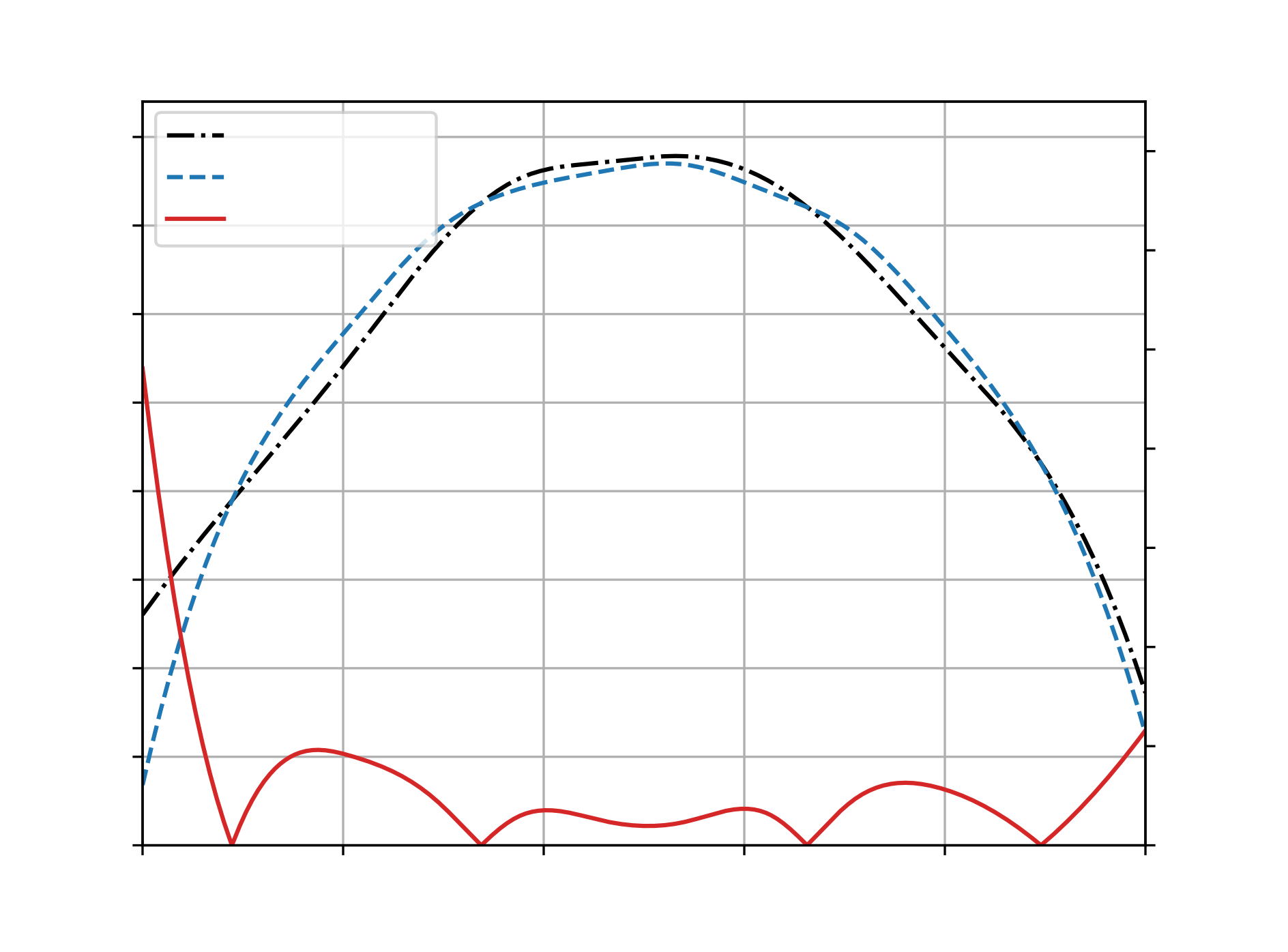}
		\put(6, 7){\tiny{\textcolor{confidence}{40}}}
		\put(6, 14){\tiny{\textcolor{confidence}{45}}}
		\put(6, 21){\tiny{\textcolor{confidence}{50}}}
		\put(6, 27.8){\tiny{\textcolor{confidence}{55}}}
		\put(6, 34.7){\tiny{\textcolor{confidence}{60}}}
		\put(6, 41.7){\tiny{\textcolor{confidence}{65}}}
		\put(6, 48.5){\tiny{\textcolor{confidence}{70}}}
		\put(6, 55.5){\tiny{\textcolor{confidence}{75}}}
		\put(6, 62.4){\tiny{\textcolor{confidence}{80}}}
		
		\put(44, 0){\scriptsize{relative $\centerx$}}
		\put(9, 4){\tiny{0.0}}
		\put(25, 4){\tiny{0.2}}
		\put(40.5, 4){\tiny{0.4}}
		\put(56, 4){\tiny{0.6}}
		\put(71.6, 4){\tiny{0.8}}
		\put(87.3, 4){\tiny{1.0}}
		
		\put(95, 29){\rotatebox{90}{\scriptsize{\textcolor{ece}{ECE [\%]}}}}
		\put(92, 7){\tiny{\textcolor{ece}{0}}}
		\put(92, 14.8){\tiny{\textcolor{ece}{2}}}
		\put(92, 22.5){\tiny{\textcolor{ece}{4}}}
		\put(92, 30.3){\tiny{\textcolor{ece}{6}}}
		\put(92, 38){\tiny{\textcolor{ece}{8}}}
		\put(90.5, 45.7){\tiny{\textcolor{ece}{10}}}
		\put(90.5, 53.8){\tiny{\textcolor{ece}{12}}}
		\put(90.5, 61){\tiny{\textcolor{ece}{14}}}
		
		\put(19, 62.3){\tiny{Precision}}
		\put(19, 59){\tiny{Confidence}}
		\put(19, 55.7){\tiny{ECE}}
	\end{overpic}
	\caption{a) Consider a miscalibrated detector where the observed precision and the confidence score decrease closer to the image boundaries. Furthermore, assume a strong positive correlation with regard to $\centerx$ and $\centery$ (top). Due to different slopes, precision and confidence exhibit a strong deviation at the boundaries and thus we have a high expected calibration error (ECE); especially in that border area (bottom). b) If we now apply a standard calibration map that only depends on the confidence score, we basically shift or in this case decrease the confidence curve (bottom). In other words, we optimize the average value of the ECE over all image locations, but don't influence the shape of the confidence curve. c) The proposed box-sensitive or multivariate calibration has access to location information and is capable of depicting dependencies between the confidence score, $\centerx$ and $\centery$. In this way, it is possible to adapt the confidence score across all locations differently.}
	\label{img:example:artificial}
\end{figure*}

Our goal is to develop calibration methods for detectors that can be applied after the training independently of the underlying detector architecture (black-box). For simple classification tasks, a confidence map $\calmodel$ is applied on top of a miscalibrated scoring classifier $\pred = \model(\singleinput)$ to deliver a calibrated confidence score $\calibrated=\calmodel(\model(\singleinput))$. The key ingredient that can be used to improve the calibration of a detector is that we have additional information about the estimated bounding box $\predbbox$. Therefore, the calibration map is not only a function of the confidence score, but also of $\predbbox$.  To define a general calibration map for binary problems, we use the logistic function and the combined input $\collect=(\pred, \predbbox)$ of size $\numcollectedvariates$ by
\begin{align}
\label{eq:calibrated_prob}
\calmodel(\collect) = \frac{1}{1 + e^{-\logit(\collect)}} .
\end{align}
Similar to the work presented in \cite{Kull2017} regarding beta-calibration, we interpret the logit $z$ as the logarithm of the posterior odds 
\begin{align}
\label{eq:logit}
\logit(\collect) &= \log \frac{\pdf(\matchedvariate = 1 | \collect)}{\pdf(\matchedvariate = 0 | \collect)} 
\approx \log \frac{\pdf(\collect | \matchedvariate = 1)}{\pdf(\collect | \matchedvariate = 0)}=\loglikelihoodratio(\collect).
\end{align}
For simplicity, we assume a uniform prior $\pdf(\matchedvariate)$ and use the approximation which in turn is the log-likelihood ratio $\loglikelihoodratio(\collect)$. In contrast to classifier calibration, we can use multivariate probability density functions to model $\pdf(\collect | \matchedvariate )$ and therefore include the bounding box information. Even though it is possible to evaluate the distribution parameters directly, we use a discriminative approach to obtain a better calibration mapping \cite{Ng2002}. Thus, the calibration process can then be posed as an optimization of a logistic regression problem. Nevertheless, using different models $\pdf(\collect | \matchedvariate )$ results in an extremely varying expressiveness of the calibration mapping, as we explain in Fig.~\ref{img:example:artificial}. 

\subsection{Conditional Independent Calibration}
\label{section:independence}
The  log-likelihood ratio in Eq.~\ref{eq:logit}, that is used to get a calibrated confidence estimate, is based on two multivariate probability density functions. If we assume conditional independence of all random variables $\collectvariate_\indexcollectedvariates$ contained in $\collectvariate$, we can simplify the joint density and the log-likelihood ratio can be expressed as
\begin{align}
\loglikelihoodratio(\collect) &= \sum_{\indexcollectedvariates=1}^\numcollectedvariates \log  \frac{\pdf(\collect_\indexcollectedvariates | \matchedvariate = 1)}{\pdf(\collect_\indexcollectedvariates | \matchedvariate = 0)} .
\end{align}
This assumptions allows multiple univariate density functions to be combined in an easy fashion. However, keeping track of dependencies and correlations between the variables is not possible. In short, dependencies between confidence, location and scale parameters are neglected.\\

\textbf{Independent Logistic Calibration}: The well-known logistic calibration or Platt scaling \cite{Platt1999} used for classification calibration is one of the simplest calibration methods. The logits of a neural network are scaled and shifted to obtain a calibrated confidence estimate. Logistic calibration is obtained using univariate normal distributions for $\calmodel(\collect)$ \cite{Kull2017}. If we assume multiple univariate normal distributions for each box quantity with equal variance, we can simply extend this method to object detection by extending the scale factor to a scale vector $\scalevec \in \realdigits^{\numcollectedvariates}$. However, instead of using the uncalibrated confidence estimate $\confidence$, we use the logit of the network as part of $\collect$ to be conform with the original formulation in \cite{Platt1999} and \cite{Kull2017}. If we assume equal variances for both normal distributions, the log-likelihood ratio of the independent logistic calibration for object detection is given by
\begin{align}
\label{eq:platt_independent}
\loglikelihoodratio_{\text{LC}}(\collect) = \collect^T \scalevec + \scalebias ,
\end{align}
with the bias $\scalebias \in \realdigits$. This method needs $\numcollectedvariates + 1$ parameters to compute a calibration mapping.\\

\textbf{Independent Beta Calibration}: In standard beta calibration \cite{Kull2017} the confidence score is assumed to follow a beta distribution in order to get a better fit in the interval $[0,1]$. Similar to Platt scaling, the beta calibration method can also be extended to the object detection task by introducing beta distributions for each box quantity $\collectvariate_\indexcollectedvariates$. Then the log-likelihood ratio can be expressed as
\begin{align}
\loglikelihoodratio_{\text{BC}}(\collect) =& \sum_{\indexcollectedvariates=1}^\numcollectedvariates \log \Bigg( \frac{\text{Beta}(\collect_\indexcollectedvariates|\betafunca^+_\indexcollectedvariates,\betafuncb^+_\indexcollectedvariates)}{\text{Beta}(\collect_\indexcollectedvariates|\betafunca^-_\indexcollectedvariates,\betafuncb^-_\indexcollectedvariates)} \Bigg) ,
\end{align}
with the beta distribution $\text{Beta}(\collect_\indexcollectedvariates|\betafunca,\betafuncb)$. The shape parameters for $\matchedvariate=1$ and $\matchedvariate=0$ are denoted by $\betafunca^+_\indexcollectedvariates$, $\betafuncb^+_\indexcollectedvariates$ and $\betafunca^-_\indexcollectedvariates$, $\betafuncb^-_\indexcollectedvariates$ respectively. We can reparametrize this expression by using
\begin{align}
\loglikelihoodratio_{\text{BC}}(\collect) = \betaparamc + \sum_{\indexcollectedvariates=1}^\numcollectedvariates \betaparama_\indexcollectedvariates \log(\collect_\indexcollectedvariates) - \betaparamb_\indexcollectedvariates \log(1 - \collect_\indexcollectedvariates) ,
\end{align}
where the distribution parameters are summarized by 
$\betaparama_\indexcollectedvariates = \betafunca^+_\indexcollectedvariates - \betafunca^-_\indexcollectedvariates$, $\betaparamb_\indexcollectedvariates = \betafuncb^-_\indexcollectedvariates - \betafuncb^+_\indexcollectedvariates$ and $\betaparamc = \sum_{\indexcollectedvariates} \log  \betafunc(\betafunca^-_\indexcollectedvariates, \betafuncb^-_\indexcollectedvariates)- \log \betafunc(\betafunca^+_\indexcollectedvariates, \betafuncb^+_\indexcollectedvariates)$. The beta function $\betafunc$ is a normalization constant. 
The reparametrization allows for standard optimization methods to determine $\calmodel(\collect)$ (cp. \cite{Kull2017}). The monotony for the confidence dimension is kept by restricting the parameters $\betaparama_1, \betaparamb_1 > 0$ as proposed in \cite{Kull2017}. In summary, this model needs $2\numcollectedvariates + 1$ parameters.

\subsection{Conditional Dependent Calibration}
As demonstrated in Fig.~\ref{img:example:artificial}, it can be beneficial to also capture correlations between the confidence and the bounding box quantities. If we keep the original formulation from Eq.~\ref{eq:logit}, we can use multivariate probability density functions to define a calibration map while keeping track of possible relationships.\\

\textbf{Dependent Logistic Calibration}: Instead of using an independent normal distribution for each quantity separately, we can also insert two multivariate normal distributions for the conditional distributions $\pdf(\collect | \matchedvariate )$ given in Eq.~\ref{eq:logit} and obtain
\begin{align}
\loglikelihoodratio_{\text{LC}}(\collect) = \frac{1}{2}\Big[(\collect_-^T \covMatP_-^{-1} \collect^-) - (\collect_+^T \covMatP_+^{-1} \collect^+)\Big] + c ,
\end{align}
with $\meanvec^+$, $\meanvec^- \in \realdigits^{\numcollectedvariates}$ and $\covMatP^+$, $\covMatP^- \in \realdigits^{\numcollectedvariates \times \numcollectedvariates}$ as the mean vectors and covariance matrices for $\matchedvariate=1$ and $\matchedvariate=0$, respectively. Furthermore, we define $c = \log \frac{|\covMatP^-|}{|\covMatP^+|}$ as well as $\collect^+ = \collect - \meanvec^+$ and $\collect^- = \collect - \meanvec^-$. To pre\-serve symmetric and positive definite covariance matrices during optimization, we estimate the entries of two matrices $\covMatCompound_+^{-1}, \covMatCompound_-^{-1} \in \realdigits^{\numcollectedvariates \times \numcollectedvariates}$ and assume $\covMatP^{-1} = (\covMatCompound^T \covMatCompound)^{-1} = \covMatCompound^{-1} (\covMatCompound^{-1})^T$. This method needs $2(\numcollectedvariates^2 + \numcollectedvariates) + 1$ parameters to be optimized. \\

\textbf{Dependent Beta Calibration}: The extension of the independent beta calibration from section \ref{section:independence} is not that straight forward to enable a tracking of possible correlations between the variables. The natural multivariate extension of a beta distribution is a Dirichlet distribution. However, the random variables represented by a Dirichlet are defined over a $\numcollectedvariates$ simplex which is not suitable for calibration. Therefore, we propose to use a generalized beta distribution for calibration established by \cite{Libby1982} and defined as
\begin{align}
\pdf(\collect | \matchedvariate) = \frac{1}{\betafunc(\betafunca_0, ..., \betafunca_\numcollectedvariates)} \frac{\prod^\numcollectedvariates_{\indexcollectedvariates=1} \Big[ \betaratio_\indexcollectedvariates^{\alpha_\indexcollectedvariates} (\collect_\indexcollectedvariates^\ast)^{\alpha_\indexcollectedvariates - 1} \big( \frac{\collect_\indexcollectedvariates^\ast}{\collect_\indexcollectedvariates} \big)^{2} \Big]}{\Big[ 1 + \sum_{\indexcollectedvariates=1}^\numcollectedvariates \betaratio_\indexcollectedvariates \collect_\indexcollectedvariates^\ast \Big]^{\sum_{\indexcollectedvariates=0}^\numcollectedvariates \betafunca_\indexcollectedvariates}} ,
\end{align}
with the shape parameters $\betafunca_\indexcollectedvariates, \betafuncb_\indexcollectedvariates$ restricted to $\betafunca_\indexcollectedvariates, \betafuncb_\indexcollectedvariates > 0 \quad \forall \indexcollectedvariates \in \{0,\ldots,\numcollectedvariates\}$. Using $\betaratio_\indexcollectedvariates = \frac{\betafuncb_\indexcollectedvariates}{\betafuncb_0}$ and $\collect_\indexcollectedvariates^\ast = \frac{\collect_\indexcollectedvariates}{1 - \collect_\indexcollectedvariates}$,
the resulting log-likelihood ratio $\loglikelihoodratio_{\text{BC}}(\collect)$ is given by 
\begin{align}
&\sum_{\indexcollectedvariates=1}^\numcollectedvariates \Bigg[ \betafunca_\indexcollectedvariates^+ \log (\betaratio_\indexcollectedvariates^+)- \betafunca_\indexcollectedvariates^- \log (\betaratio_\indexcollectedvariates^-) + ( \betafunca_\indexcollectedvariates^+ - \betafunca_\indexcollectedvariates^-) \log (\collect_\indexcollectedvariates^\ast) \Bigg] + \\ \nonumber
&\sum_{\indexcollectedvariates=0}^\numcollectedvariates \Bigg[ \betafunca_\indexcollectedvariates^- \log \Bigg( \sum_{j=1}^\numcollectedvariates \betaratio_j^- \collect_j^\ast \Bigg) - \betafunca_\indexcollectedvariates^+ \log \Bigg( \sum_{j=1}^\numcollectedvariates \betaratio_j^+ \collect_j^\ast \Bigg) \Bigg] + c ,
\end{align}
where $c = \log \betafunc(\betafunca_0^-, ..., \betafunca_\indexcollectedvariates^-)-\log\betafunc(\betafunca_0^+, ..., \betafunca_\indexcollectedvariates^+)$.
Positive correlations are captured by the distribution, whereas negative correlations are neglected \cite{Olkin2015}. 
With this formulation, we need to optimize $4(\numcollectedvariates + 1) + 1$ parameters.\\

\textbf{Multidimensional Histogram Binning}: Histogram binning is one of the simplest methods to calibrate confidence estimates. The confidence space is divided into $\numbins$ equally sized bins to group all predictions of a classifier. This allows calculating the gap between average precision and confidence in each bin. Afterwards, a confidence map $\calmodel(\collect)$ is obtained by assigning a new confidence score to each prediction that equals the observed precision in the according bin. Instead of integrating the histogram binning in our framework (Eq.~\ref{eq:calibrated_prob}), we directly use a multidimensional binning for each dimension of $\collect$. This allows a fine-grained calculation of the average precision with respect to different box information like image location, width, and height to obtain new confidence scores. The number of parameters however grows exponentially with the number of dimensions $\numbins_{\text{total}} = \prod^\numcollectedvariates_{\indexcollectedvariates=1} \numbins_\indexcollectedvariates$ and thus $\numbins_\indexcollectedvariates$ needs to be chosen carefully for each dimension.

	\section{Measuring Miscalibration of Detectors}
For classification problems, the expected calibration error (ECE) is used to measure the miscalibration of a model \cite{Naeini2015, Guo2018}. In the classification domain, this score describes the expected deviation of the observed accuracy from the estimated confidence 
\begin{align}
\expectation_{\predvariate} \Big[\big| \prob(\predclassvariate = \classvariate | \predvariate = \probscore) - \probscore \big| \Big].
\end{align}
Since $\predvariate$ is a continuous random variable, the ECE is approximated by binning the confidence space of $\predvariate$ into $\numbins$ equally spaced bins where $\bin(\indexbins)$ is the set of all samples in a single bin. The ECE is thus approximated by
\begin{align}
\label{eq:ece_classification_approximation}
\text{ECE} = \sum^\numbins_{\indexbins=1} \frac{|\bin(\indexbins)|}{|\trainingsset|} \cdot \Big|\text{acc}\left(\indexbins\right) - \text{conf}\left(\indexbins\right)\Big| ,
\end{align}
where $|\trainingsset|$ is the total amount of samples. The average accuracy and confidence in each bin is denoted by $\text{acc}(\indexbins)$ and $\text{conf}(\indexbins)$, respectively.
Intuitively, we could use this formulation to measure the miscalibration of detectors as well. The example evaluation of different methods in Fig.~\ref{img:example:artificial} (top) however shows the weakness of the standard ECE. Comparing the different heatmaps for each method reveals a significant improvement of the confidence calibration when we use additional box information instead of a global calibration. However, this fact is reflected very poorly by the standard ECE score. In this case, the global calibration (ECE $=1.1\%$) outperforms the dependent logistic calibration  (ECE $=2.2\%$) in terms of the ECE (cp. Fig.~\ref{img:example:artificial}). The ECE uses the confidence of each sample independent of the box properties to apply binning and to calculate an average precision. But for object detection tasks, the average precision may vary in different areas of the image. However, our proposed methods for location-dependent calibration adjust the confidence differently for different image regions. This cannot be captured by the standard ECE - thus, a location-dependent calibration may not lead to an improvement of the standard ECE score.

\begin{figure}[ht]
	\centering
	\begin{overpic}[width=0.48\linewidth]{img/ece_heatmap_xy_default_cleared.png}
		\put(13,88){\small{Default Calibration Error}}
		
		\put(0, 35){\rotatebox{90}{\scriptsize{relative $\centery$}}}
		\put(5.5, 8.6){\tiny{0.0}}
		\put(5.5, 42.8){\tiny{0.5}}
		\put(5.5, 76.8){\tiny{1.0}}
		
		\put(42, -1){\scriptsize{relative $\centerx$}}
		\put(10.2, 4){\tiny{0.0}}
		\put(45, 4){\tiny{0.5}}
		\put(80, 4){\tiny{1.0}}
		
		\put(82, 82){\tiny{\mbox{D-ECE [\%]}}}
		\put(93.5, 10.7){\tiny{5}}
		\put(93, 24.3){\tiny{10}}
		\put(93, 37.7){\tiny{15}}
		\put(93, 50.8){\tiny{20}}
		\put(93, 64.2){\tiny{25}}
		\put(93, 77.3){\tiny{30}}

	\end{overpic}
	\vspace{-0.5em}
	\begin{overpic}[width=0.48\linewidth]{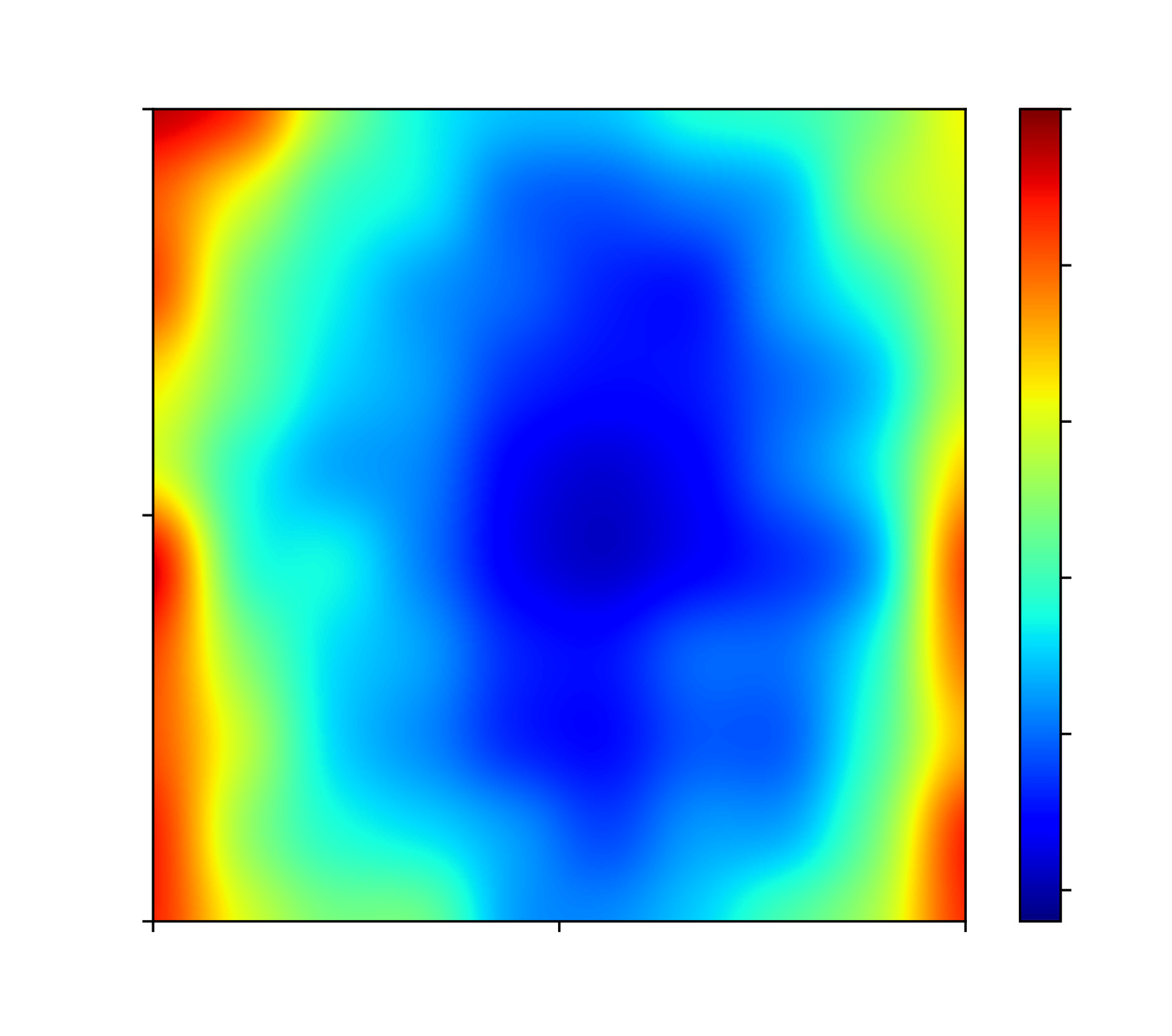}
		\put(13,88){\small{After Histogram Binning}}
		\put(5.5, 8.6){\tiny{0.0}}
		\put(5.5, 42.8){\tiny{0.5}}
		\put(5.5, 76.8){\tiny{1.0}}
		
		\put(42, -1){\scriptsize{relative $\centerx$}}
		\put(10.2, 4){\tiny{0.0}}
		\put(45, 4){\tiny{0.5}}
		\put(80, 4){\tiny{1.0}}
		
		\put(82, 82){\tiny{\mbox{D-ECE [\%]}}}
		\put(93.5, 10.7){\tiny{5}}
		\put(93, 24.3){\tiny{10}}
		\put(93, 37.7){\tiny{15}}
		\put(93, 50.8){\tiny{20}}
		\put(93, 64.2){\tiny{25}}
		\put(93, 77.3){\tiny{30}}
	\end{overpic}
	
	\vspace{0.5em}
	\begin{overpic}[width=0.48\linewidth]{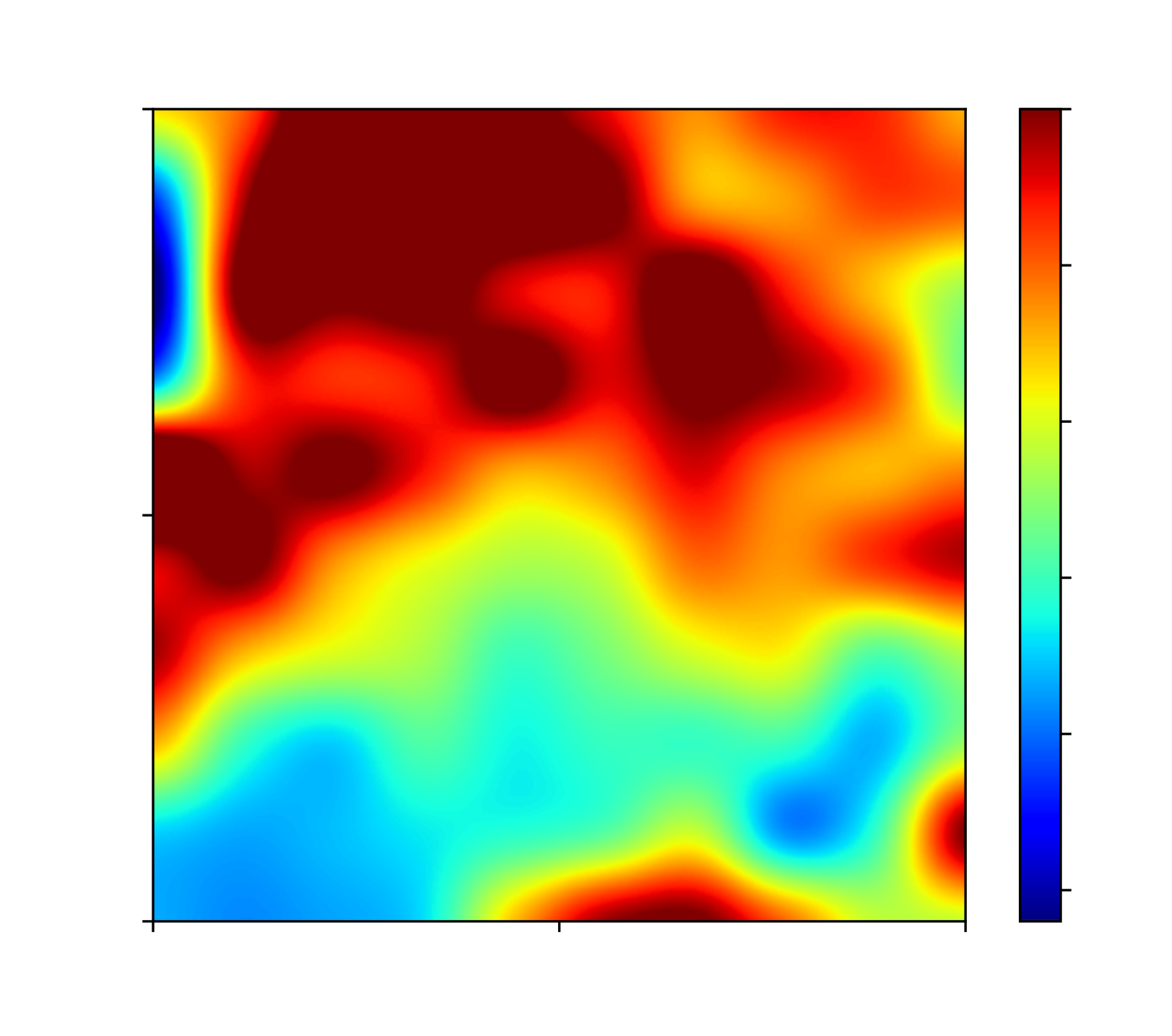}
		\put(0, 27){\rotatebox{90}{\scriptsize{relative height}}}
		\put(5.5, 8.6){\tiny{0.0}}
		\put(5.5, 42.8){\tiny{0.5}}
		\put(5.5, 76.8){\tiny{1.0}}
		
		\put(35, -1){\scriptsize{relative width}}
		\put(10.2, 4){\tiny{0.0}}
		\put(45, 4){\tiny{0.5}}
		\put(80, 4){\tiny{1.0}}
		
		\put(82, 82){\tiny{\mbox{D-ECE [\%]}}}
		\put(93.5, 10.7){\tiny{5}}
		\put(93, 24.3){\tiny{10}}
		\put(93, 37.7){\tiny{15}}
		\put(93, 50.8){\tiny{20}}
		\put(93, 64.2){\tiny{25}}
		\put(93, 77.3){\tiny{30}}
	\end{overpic}
	\vspace{0.2em}
	\begin{overpic}[width=0.48\linewidth]{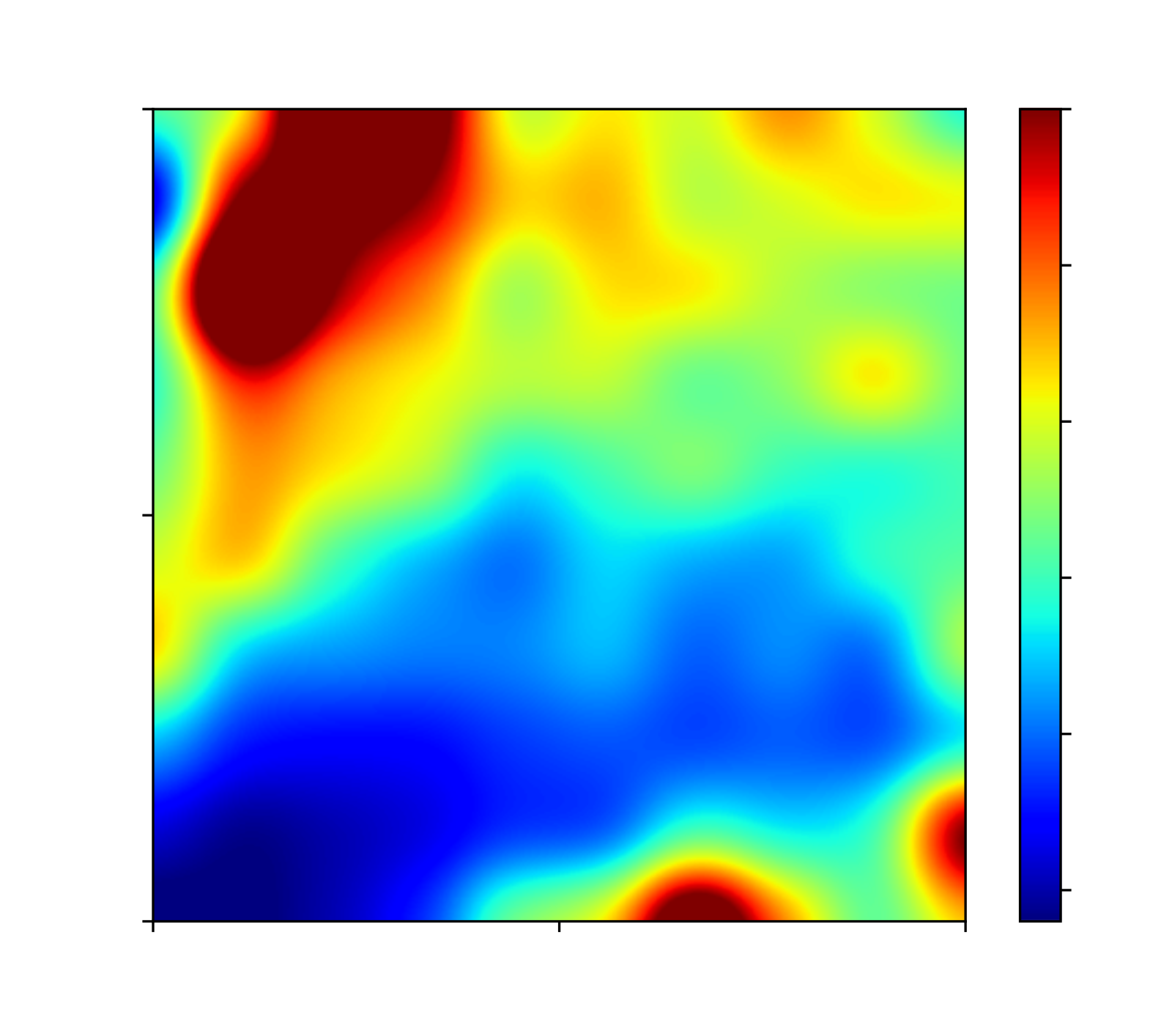}
		\put(5.5, 8.6){\tiny{0.0}}
		\put(5.5, 42.8){\tiny{0.5}}
		\put(5.5, 76.8){\tiny{1.0}}
		
		\put(35, -1){\scriptsize{relative width}}
		\put(10.2, 4){\tiny{0.0}}
		\put(45, 4){\tiny{0.5}}
		\put(80, 4){\tiny{1.0}}
		
		\put(82, 82){\tiny{\mbox{D-ECE [\%]}}}
		\put(93.5, 10.7){\tiny{5}}
		\put(93, 24.3){\tiny{10}}
		\put(93, 37.7){\tiny{15}}
		\put(93, 50.8){\tiny{20}}
		\put(93, 64.2){\tiny{25}}
		\put(93, 77.3){\tiny{30}}
	\end{overpic}
	\caption{Calibration heatmap of R-FCN \cite{Dai2016, Huang2017a} detections on COCO with IoU @ 0.6. The left column shows the baseline miscalibration measured over the relative $\centerx$ and $\centery$ coordinates (top row) and the relative width/height (bottom row) respectively. The right column shows the \mbox{D-ECE} scores after calibration with histogram binning that is also able to capture those additional box information. For the corner case, low relative width and high relative height, there are only a few data samples available. Thus, the \mbox{D-ECE} calculation is likely to be inaccurate in that area.}
	\label{img:result:heatmap}
\end{figure}

For this reason, we also integrate the box information into the ECE calculation. Therefore, we define the detection ECE (D-ECE) as the expected deviation of the observed precision with respect to the given box properties. Using Eq.~\ref{eq:calibration} we get
\begin{align}
\label{eq:ece_detection}
& \expectation_{\predvariate, \predbboxvariate} \Big[ \big| \prob(\matchedvariate = 1 | \predvariate = \probscore, \predclassvariate = \class, \predbboxvariate = \bbox) - \probscore \big| \Big]  .
\end{align}
Thus, the \mbox{D-ECE} for a given class $\class$ depends on the IoU and on the number of additional box information. If we calculate the expectation with respect to a subset of the variables, e.g. $\predvariate$, calibration heatmaps can be generated (cp. Fig.~\ref{img:result:heatmap}). 

Similar to Eq.~\ref{eq:ece_classification_approximation}, we calculate the multidimensional \mbox{D-ECE} by partitioning the confidence space as well as the box property space in each dimension $\indexcollectedvariates$ into $\numbins_\indexcollectedvariates$ equally spaced bins. If we iterate over all bins in all dimensions, the total amount of bins is given by $\numbins_{\text{total}} = \prod^\numcollectedvariates_{\indexcollectedvariates=1} \numbins_\indexcollectedvariates$ and the \mbox{D-ECE} is thus given by
\begin{align}
\text{D-ECE}_\numcollectedvariates = \sum^{\numbins_\text{total}}_{\indexbins=1} \frac{|\bin(\indexbins)|}{|\trainingsset|} \cdot \Big|\text{prec}\left(\indexbins\right) - \text{conf}\left(\indexbins\right)\Big| ,
\end{align}
where $\text{prec}(\indexbins)$ and $\text{conf}(\indexbins)$ denote the average precision and confidence in each bin, respectively.

On the one hand, a higher number of bins leads to a more accurate evaluation of the D-ECE score, especially if the data density is a very steep or quickly changing function. On the other hand, the number of samples per bin decreases as we increase the number of bins. Thus, the estimation of the average precision and confidence in each bin is less accurate. Furthermore, due to the non-linearity and bin migration of samples, the D-ECE scores with different number of dimensions and/or bins cannot be compared directly to each other. Therefore, we suggest to evaluate calibration methods for object detection with a D-ECE of at least the same dimensionality as was used for calibration. 

Consider the following example: Method A uses only the confidences for calibration without any additional box information ($\numcollectedvariates=1$) while method B also takes the $\centerx$ position of the bounding boxes into account ($\numcollectedvariates=2$). Method C calibrates with respect to all three quantities, confidence $\pred$ and $\centerx$, $\centery$ position ($\numcollectedvariates=3$).
To compare these models, we need the D-ECE$_\numcollectedvariates$ ($\numcollectedvariates=3$ with $\centerx$, $\centery$ position) because C uses the confidence and 2 additional box quantities for calibration. A comparison with the standard ECE is not reasonable. Note that calibration methods with other box information (e.g. width or height) should not be compared to B or C.

	\section{Experiments}
\begin{table*}[h]
	\centering
	\begin{subtable}{0.33\textwidth}
		\centering
		\resizebox{0.95\textwidth}{!}{%
			\begin{tabular}{llrrrr}
\toprule
&      &  $(\pred)$ &  $(\pred, \centerx, \centery)$ &  $(\pred, \width, \height)$ &  full \\
\midrule
\multicolumn{2}{l}{baseline} & 19.235 & 18.868 & 18.265 & 15.255 \\
\midrule
\multirow{2}{*}{HB} & conf. &   \textbf{0.890} &  5.121 &  5.181 &  5.790 \\
   & dep. &       - &  \textbf{4.557} &  \textbf{3.739} &  5.727 \\
\cline{1-6}
\multirow{3}{*}{LC} & conf. &   1.282 &  5.244 &  5.321 &  5.752 \\
   & indep. &       - &  5.283 &  4.300 &  5.011 \\
   & dep. &       - &  5.310 &  4.118 &  \textbf{4.935} \\
\cline{1-6}
\multirow{3}{*}{BC} & conf. &   1.321 &  5.250 &  5.283 &  5.751 \\
   & indep. &       - &  5.296 &  4.557 &  5.097 \\
   & dep. &       - &  5.290 &  4.078 &  5.105 \\
\bottomrule
\end{tabular}

		}
		\caption{F. R-CNN ResNet-50 \cite{Ren2015}, IoU@0.6.}
	\end{subtable}
	\begin{subtable}{0.33\textwidth}
		\centering
		\resizebox{0.925\textwidth}{!}{%
			\begin{tabular}{llrrrr}
\toprule
&      &  $(\pred)$ &  $(\pred, \centerx, \centery)$ &  $(\pred, \width, \height)$ &  full \\
\midrule
\multicolumn{2}{l}{baseline} &  11.675 & 11.623 & 10.490 & 8.781 \\
\midrule
\multirow{2}{*}{HB} & conf. &   \textbf{1.178} &  5.392 &  4.332 & 4.930 \\
   & dep. &       - &  \textbf{4.991} &  \textbf{3.360} & 4.975 \\
\cline{1-6}
\multirow{3}{*}{LC} & conf. &   1.232 &  5.404 &  4.437 & 4.891 \\
   & indep. &       - &  5.412 &  3.387 & 4.298 \\
   & dep. &       - &  5.460 &  3.419 & \textbf{4.181} \\
\cline{1-6}
\multirow{3}{*}{BC} & conf. &   1.299 &  5.431 &  4.450 & 4.917 \\
   & indep. &       - &  5.460 &  3.646 & 4.237 \\
   & dep. &       - &  5.412 &  3.422 & 4.369 \\
\bottomrule
\end{tabular}

		}	
		\caption{R-FCN ResNet-101 \cite{Dai2016}, IoU@0.6.}
	\end{subtable}
	\begin{subtable}{0.33\textwidth}
		\centering
		\resizebox{0.94\textwidth}{!}{%
			\begin{tabular}{llrrrr}
\toprule
&      &  $(\pred)$ &  $(\pred, \centerx, \centery)$ &  $(\pred, \width, \height)$ &  full \\
\midrule
\multicolumn{2}{l}{baseline} &   6.648 &  8.286 & 11.400 & 10.239 \\
\midrule
\multirow{2}{*}{HB} & conf. &   \textbf{1.278} &  5.766 & 10.723 &  7.429 \\
   & dep. &       - &  \textbf{5.615} &  6.544 &  7.716 \\
\cline{1-6}
\multirow{3}{*}{LC} & conf. &   1.306 &  6.085 & 10.707 &  7.575 \\
   & indep. &       - &  6.103 &  8.356 &  5.763 \\
   & dep. &       - &  6.017 &  5.638 &  5.111 \\
\cline{1-6}
\multirow{3}{*}{BC} & conf. &   1.297 &  6.016 & 10.721 &  7.533 \\
   & indep. &       - &  6.036 &  5.265 &  5.090 \\
   & dep. &       - &  6.041 &  \textbf{4.715} &  \textbf{4.916} \\
\bottomrule
\end{tabular}

		}	
		\caption{SSD Inception v2 \cite{Liu2016}, IoU@0.6.}
	\end{subtable}
	\vspace{1em}
	
	\begin{subtable}{0.33\textwidth}
		\centering
		\resizebox{0.95\textwidth}{!}{%
			\begin{tabular}{llrrrr}
\toprule
&      &  $(\pred)$ &  $(\pred, \centerx, \centery)$ &  $(\pred, \width, \height)$ &  full \\
\midrule
\multicolumn{2}{l}{baseline} &  32.458 & 31.805 & 30.990 & 26.468 \\
\midrule
\multirow{2}{*}{HB} & conf. &   \textbf{0.910} &  4.987 &  5.678 &  5.821 \\
   & dep. &       - &  \textbf{4.392} &  \textbf{4.532} &  6.340 \\
\cline{1-6}
\multirow{3}{*}{LC} & conf. &   1.667 &  5.321 &  5.603 &  5.290 \\
   & indep. &       - &  5.279 &  5.419 &  4.883 \\
   & dep. &       - &  5.196 &  4.776 &  \textbf{4.755} \\
\cline{1-6}
\multirow{3}{*}{BC} & conf. &   1.266 &  5.367 &  5.370 &  5.201 \\
   & indep. &       - &  5.214 &  5.485 &  5.047 \\
   & dep. &       - &  5.327 &  4.733 &  4.920 \\
\bottomrule
\end{tabular}

		}
		\caption{F. R-CNN ResNet-50 \cite{Ren2015}, IoU@0.75.}
	\end{subtable}
	\begin{subtable}{0.33\textwidth}
		\centering
		\resizebox{0.95\textwidth}{!}{%
			\begin{tabular}{llrrrr}
\toprule
&      &  $(\pred)$ &  $(\pred, \centerx, \centery)$ &  $(\pred, \width, \height)$ &  full \\
\midrule
\multicolumn{2}{l}{baseline} &  26.087 & 25.207 & 23.862 & 20.394 \\
\midrule
\multirow{2}{*}{HB} & conf. &  \textbf{1.014} &  5.279 &  4.583 &  5.343 \\
   & dep. &       - &  \textbf{5.005} &  4.171 &  5.967 \\
\cline{1-6}
\multirow{3}{*}{LC} & conf. &   1.679 &  5.769 &  4.231 &  4.849 \\
   & indep. &       - &  5.768 &  4.099 &  4.740 \\
   & dep. &       - &  5.609 &  3.976 &  \textbf{4.734} \\
\cline{1-6}
\multirow{3}{*}{BC} & conf. &   1.485 &  5.769 &  4.188 &  4.785 \\
   & indep. &       - &  5.604 &  4.144 &  4.778 \\
   & dep. &       - &  5.756 &  \textbf{3.872} &  4.772 \\
\bottomrule
\end{tabular}

		}	
		\caption{R-FCN ResNet-101 \cite{Dai2016}, IoU@0.75.}
	\end{subtable}
	\begin{subtable}{0.33\textwidth}
		\centering
		\resizebox{0.95\textwidth}{!}{%
			\begin{tabular}{llrrrr}
\toprule
&      &  $(\pred)$ &  $(\pred, \centerx, \centery)$ &  $(\pred, \width, \height)$ &  full \\
\midrule
\multicolumn{2}{l}{baseline} &  19.668 & 18.761 & 19.470 & 18.067 \\
\midrule
\multirow{2}{*}{HB} & conf. &   \textbf{1.262} &  5.654 & 10.772 &  7.581 \\
   & dep. &       - &  \textbf{5.383} &  5.846 &  7.718 \\
\cline{1-6}
\multirow{3}{*}{LC} & conf. &   1.688 &  5.888 & 10.599 &  7.308 \\
   & indep. &       - &  5.872 &  7.700 &  4.958 \\
   & dep. &       - &  5.865 &  5.541 &  \textbf{4.130} \\
\cline{1-6}
\multirow{3}{*}{BC} & conf. &   1.796 &  6.002 & 10.649 &  7.366 \\
   & indep. &       - &  5.923 &  4.839 &  4.140 \\
   & dep. &       - &  5.893 &  \textbf{4.713} &  4.184 \\
\bottomrule
\end{tabular}

		}	
		\caption{SSD Inception v2 \cite{Liu2016}, IoU@0.75.}
	\end{subtable}
	\caption{Calibration scores for different networks. The baseline is the default miscalibration of the network. Each column shows the \mbox{D-ECE} score for a specific dimensionality, that is determined by the additional bounding box information  (global or confidence only, \textit{$(\pred, \centerx, \centery)$}, \textit{$(\pred, \width, \height)$} or full). Each row contains the evaluation of a calibration method (e.g. histogram binning (HB), logistic (LC) or beta (BC) calibration). Each method is subdivided into the categories 1) confidence only (conf.), 2) box information assumed to be independent (indep.) or 3) box information assumed to be dependent (dep.). The best results are mainly achieved by multidimensional histogram binning for confidence only calibration, whereas multivariate logistic and beta calibration outperform the histogram binning in the box-sensitive case.}
	\label{table:calibration_results}
\end{table*}
\subsection{Evaluation Protocol}
We apply our extended calibration methods to several pretrained object detectors available in \cite{Huang2017a}. All networks use a non-max suppression with a probability threshold of $0.3$ and an IoU threshold of $0.6$. We use the COCO validation dataset 2017 \cite{Lin2014} with images licensed for commercial use only to demonstrate the effectiveness of our approach. Unfortunately, we cannot use the according COCO test set as there are no annotations publicly available. Thus, we are using a random split of the validation set and use 70\% for building the calibration scheme and the remaining images for testing. The random split is repeated 20 times to obtain an average calibration result for each method. 

In general, one should use all influential factors $(c_x, c_y, w, h)$ for measuring the D-ECE metric and for box-sensitive confidence calibration. However, for anchor-free methods \cite{Zhu2019} or semantic segmentation only the position is available. Furthermore, for safety critical applications the calibration goal depends on the assurance case which could be independent of the scale. Thus, we evaluated several setups to demonstrate our generic approach.
For the one-dimensional calibration case (confidence only), we use 15 bins for the histogram binning and 20 bins for the computation of the ECE. For $K=3$ ($(\pred, \centerx, \centery)$ or $(\pred, \width, \height)$), we use $\numbins_\indexcollectedvariates = 5$ for histogram-based calibration and $\numbins_\indexcollectedvariates = 8$ for the D-ECE. Finally, in the complete box-sensitive calibration case $K=5$, the histogram calibration is performed with $\numbins_\indexcollectedvariates = 3$ bins in each dimension while using a binning of $\numbins_\indexcollectedvariates = 5$ for the D-ECE. The robustness of the D-ECE calculation is increased by neglecting bins with less than 8 samples.

\subsection{Results}

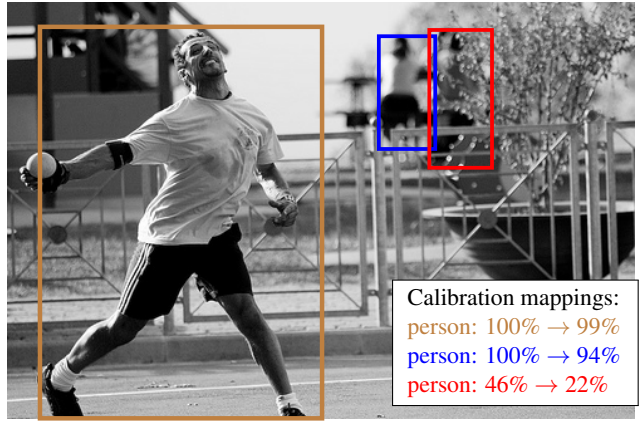
\begin{figure}[h]
	\begin{center}
		\begin{overpic}[width=1.0\linewidth]{img/000000515579.jpg}
			
			\linethickness{0.6mm} \color{brown}
			\put(0, 0){\polygon(2, 0)(2, 62.5)(47,62.5)(47,0)}
			
			\linethickness{0.5mm} \color{blue}
			\put(0, 0){\polygon(55, 43)(55, 61)(64,61)(64,43)}
			
			\linethickness{0.5mm} \color{red}
			\put(0, 0){\polygon(62, 40)(62, 62)(72, 62)(72,40)}
			
			\put(55, 11){
				\fcolorbox{black}{white}{
					\begin{minipage}[l]{8.4em}
						\small 
						{\color{black}Calibration mappings:}\\
						{\color{brown} person: 100\% $\rightarrow$ 99\%}\\
						{\color{blue} person: 100\% $\rightarrow$ 94\%}\\
						{\color{red} person: 46\% $\rightarrow$ 22\%}
						
				\end{minipage}}
			}
		\end{overpic}
	\end{center}
	\caption{Qualitative R-FCN calibration results for predictions on COCO with equal confidence but with different po\-si\-tions and scales. Using the dependent logistic calibration, the confidence of the brown detection is slightly adapted, whereas the remaining detections undergo major changes to indicate the uncertainty for predictions with those confidences, positions, and scales.}
	\label{img:qualitative_result}
\end{figure}

A baseline is obtained by evaluating each detector without calibration, but with D-ECE measures of different dimensionality. The results are presented in the first row of Table~\ref{table:calibration_results}. In the first experiment, we apply global calibration methods without any additional box information. The results for histogram binning, logistic and beta calibration are presented in the first column of Table~\ref{table:calibration_results}. In contrast to the observations for classification presented in \cite{Guo2018}, the simple histogram binning outperforms the logistic and the beta calibration in all cases. In the next two experiments (second and third column), additional bounding box quantities like position or scale are evaluated separately. Finally, all bounding box quantities are used for calibration (fourth column). As proposed in the previous section, the computation of the \mbox{D-ECE} is always chosen according to the dimensionality. For the evaluation, we applied the same IoU score of 0.6 as used for the non-max suppression of the networks. In addition, we also measured the miscalibration for a higher IoU score of 0.75. The results for different network architectures and different IoU scores are shown in Table~\ref{table:calibration_results} with the best \mbox{D-ECE} scores highlighted. Fig.~\ref{img:result:heatmap} illustrates the improved calibration in terms of a heatmap. The effect of box-sensitive confidence calibration is also qualitatively demonstrated in Fig.~\ref{img:qualitative_result}.

The results reveal that the simple histogram binning achieves the lowest calibration error for the global calibration with confidence only and fair results for calibration with respect to either $(\pred, \centerx, \centery)$ or $(\pred, \width, \height)$. For the confidence calibration with all box information available, the multivariate logistic or the beta calibration outperform the histogram binning, especially in the dependent case. Thus, it is preferable to keep track of correlations between the box quantities in order to improve the \mbox{D-ECE} score, especially for high dimensional calibration mappings. In general, the histogram binning has offered a high representational power for detection calibration but in the case of high dimensionality it is outperformed by the other methods. It should be clear that an increasing number of bins makes the histogram binning prone to overfitting and, in addition, the amount of bins grows exponentially with the number of dimensions. A representative histogram-based calibration mapping requires a dataset that compensates for the exponential bin growth.
Therefore, we conclude that either the logistic calibration or beta calibration should be preferred for high dimensional calibration mappings with only a small or medium amount of data available, whereas the histogram binning might be used for low-dimensional cases or on large datasets. 

	\section{Conclusion}

In this paper we examined how to extend common calibration methods to the object detection task. We claim that the calibration depends on the regression output of the detector, which is conform with our empirical observations on the COCO dataset. Therefore, we  propose a novel framework to measure miscalibration and to calibrate object detection models. To reveal and overcome the limitations of traditional methods, a location and scale dependent treatment of measuring miscalibration is needed. The same holds for the calibration methodology. Using the proposed framework, we are able to outperform state-of-the-art calibration methods by including additional bounding box information from the detector. The proposed framework is also useful to measure the miscalibration with respect to the box regression output. We note that this new measurement provides insights into the properties of a detector that are not revealed by any traditional evaluation metric. We have shown that multidimensional histogram binning is a good choice for low dimensional calibration mappings. The multivariate logistic or beta calibration achieve superior results for high dimensional mappings using considerably fewer parameters.

\ifcvprfinal
\section*{Acknowledgments}
This research is funded by Visteon Electronics GmbH, Kerpen, Germany and the German Federal Ministry for Economic Affairs and Energy within the project 'KI Absicherung - Safe AI for automated driving'. The authors would like to thank Visteon for their support and namely Andreas Wedel for the fruitful discussions.
\fi
	
	\newpage
	
	{\small
		\bibliographystyle{ieee_fullname}
		\bibliography{bib/bibliography}
	}
	
\end{document}